%% file: arxiv_Galilean_20261116.tex
\documentclass{article}
\usepackage{arxiv}

\usepackage{amsmath,amsfonts,amssymb}
\usepackage{bm}
\usepackage{xcolor}
\usepackage{array}
\usepackage[caption=false,font=normalsize,labelfont=sf,textfont=sf]{subfig}
\usepackage{textcomp}
\usepackage{stfloats}
\usepackage{url}
\usepackage{verbatim}
\usepackage{graphicx}
\usepackage{cite}
\usepackage{cancel}
\usepackage{tikz}
\usepackage{tikz-3dplot}
\usetikzlibrary{%
  calc,
  arrows.meta,
  positioning,
  shapes.arrows,
  decorations.pathmorphing%
}

\input{preamble_arXiv.tex}

\newcommand{\bv}{b}

\input{notation_defs.tex}

\newcommand{\publicationdetails}
{to appear in Robotics and Automation Magazine}
\newcommand{\publicationversion}
{Preprint}

\begin{document}

\title{Galilean Symmetry in Robotics}
\headertitle{Galilean Symmetry in Robotics}

\author{
\href{https://orcid.org/0000-0002-7803-2868}{\includegraphics[scale=0.06]{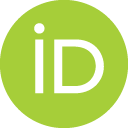}\hspace{1mm}
Robert Mahony}
\\
	Systems Theory and Robotics Group, \\
	Australian National University, \\
    Australia \\
	\texttt{Robert.Mahony@anu.edu.au} \\
	\And	\href{https://orcid.org/0000-0002-5528-6136}{\includegraphics[scale=0.06]{orcid.png}\hspace{1mm}
Jonathan Kelly}
\\
Space and Terrestrial Autonomous Robotics Systems Laboratory, \\
University of Toronto \\
Canada. \\
	\texttt{jonathan.kelly@robotics.utias.utoronto.ca} \\
	\And	\href{https://orcid.org/0000-0001-6906-5409}{\includegraphics[scale=0.06]{orcid.png}\hspace{1mm}
Stephan Weiss}
\\
Control of Networked Systems group,  \\
University of Klagenfurt,\\
Austria. \\
	\texttt{stephan.weiss@aau.at} \\
}

\maketitle


\begin{abstract}
In this paper, we present a robotics-tailored exposition of Galilean symmetry that leverages the community's familiarity with and understanding of rigid-body transformations and pose representations.
Our approach contrasts with common treatments in the physics literature that introduce Galilean symmetry as a stepping stone to Einstein's relativity.
We show that the classical homogeneous matrix algebra known well to roboticist translates directly into the Galilean modeling framework and provides a powerful tool for tackling robotics problems involving inertial measurements and spatio-temporal variables. 
We demonstrate the potential of the approach by developing a powerful model for studying motion of vehicles in an Earth fixed rotating reference frame, and developing a fusion algorithm for measurements with uncertain time-stamps. 
\end{abstract}


\section{Introduction}
\label{sec:intro_PartI}

Galilean symmetry is the natural symmetry of inertial motion that underpins Newtonian physics.
Although rigid-body symmetry is one of the most established and fundamental tools in robotics, there appears to be no comparable treatment of Galilean symmetry for a robotics audience.
We believe that the time is right for the robotics community to benefit from rediscovering and extending this classical material and applying it to modern problems.

The starting point for a discussion of Galilean space-time is to understand what is meant by the term ``inertial''. 
An inertial frame is a non-accelerating, non-rotating, frame in the absence of gravity.
Any frame moving at a constant velocity relative to an inertial frame is also inertial and the concept is intrinsic, that is independent of which frame is used as reference. 
Inertial reference frames provide a consistent `stage' for describing motion---physical laws are \emph{invariant} in the sense that they retain the same form regardless of an observer's inertial frame of reference. 
This invariance serves as the foundation for Galilean relativity, where space and time are unified into a four-dimensional manifold of spatio-temporal \textit{events} that together form Galilean spacetime.
Relationships between inertial frames, and between events observed in different inertial frames, are described by Galilean transformations, which include translations in space and time, rotations of spatial coordinates, velocity boosts, and time offsets \cite{2011_Holm_Geometric_Part_II}.
The set of Galilean transformations forms a 10-dimensional Lie group, called the Galilean group\footnote{
The Poincar\'{e} group is more general than the Galilean group, as Poincar\'{e} transformations account for the finite speed of light and for relativistic effects. 
However, Galilean transformations are sufficient for almost all (if not all) robotics applications.
} 
and denoted\footnote{
The `special' in \emph{special Euclidean} $\SE(3)$ and \emph{special orthogonal} $\SO(3)$ groups refers to the exclusion of reflection symmetries that can act on general Euclidean space. 
Galilean space-time does not admit reflection symmetry by construction and we refer to the group of its symmetries simply as the \emph{Galilean group} with notation $\Gal(3)$.}
$\Gal(3)$.
Notably, many of the classical groups that are frequently applied in robotics problems are proper subgroups of the Galilean group, including the special orthogonal group, $\SO(3)$, the special Euclidean group, $\SE(3)$, and the group of extended poses, $\SE_{2}(3)$ \cite{2020_Barrau_Mathematical,brossard2021associating}.
However, $\Gal(3)$ has the key additional property that it acts on events (with both space and time coordinates) rather than just modelling spatial variables. 

Perhaps surprisingly, despite being well known in the geometric mechanics literature (e.g., discussed in \cite{1999_Marsden_Introduction} and other foundational texts), the Galilean group has received little attention in robotics.
To the best of the authors' knowledge, the first robotics-specific work to explicitly consider Galilean symmetry is \cite{2021_Giefer_Uncertainties}, but for $\Gal(2)$ only.
However, Galilean symmetry is the natural symmetry of inertial sensing and, without formal recognition, it has underpinned significant advances in Inertial Navigation Systems over the last ten years \cite{barrau2016invariant}, \cite{Barrau_2023},\cite{FornasierEquivariantSymmetries2025}, \cite{2024_Ge_LCSS}, \cite{2023_vanGoor_EqF}. 
The original work on velocity aided attitude \cite{2009_bonnable_vaa} has a group structure corresponding to the homogeneous Galilean group. 
The  \emph{double direct spatial isometry} group identified by Barrau \etal \cite{barrau2014invariant,barrau2016invariant,Barrau_2023} is the \emph{isochronous Galilean} group \cite{levy1971galilei} that is now commonly known as the \emph{extended pose} group \cite{2020_Barrau_Mathematical,brossard2021associating}, in the robotics literature.
Galilean symmetry is a core technology in recent papers on pre-integration \cite{2019_Fourmy_Absolute,2024_Shalaby_Multi-robot,Delama_2024} and time-lag estimation \cite{2026_Delama}.
Mahony \cite{2024_Mahony_talk} used Galilean symmetry to model kinematics for robotic systems in rotating frames of reference. 
Recent work in robotics community \cite{barfoot2025integral,2024_Kelly_arxiv,Delama_2024} shows growing awareness of the Galilean group and the importance of understanding and exploiting this modelling framework in robotics problems. 

Given recent interest, our goal in this tutorial is to provide an introductory treatment of Galilean symmetry and the associated matrix Lie-group algebra tailored for the robotics audience.
We describe the structure and composition of the Galilean group, highlighting its relevance for robotics applications.
We show that the homogeneous matrix algebra that has been so powerful in rigid-body modelling in robotics can be directly extended to Galilean frames providing roboticists with a ready made intuition for working with this new framework, and highlight what additional information is available, and where the pitfalls are that may catch out the unwary. 
We consider two example problems that are well established in the robotics and aerospace literature and demonstrate how the Galilean framework can be used to provide significant insight. 
The first problem we consider is modelling motion of a robotic vehicle in an Earth fixed rotating frame with gravity. 
The \emph{Galilean frame} formulation leads to a clean and intuitive derivation of the kinematic equations of motion. 
A key contribution is the careful discussion of the various gravitational and centripetal acceleration terms present in such models. 
The second example considered is sensor fusion with temporal uncertainty, where the Galilean matrix state representation allows one to model asynchronous, delayed, and noisy measurements in an integrated and consistent manner. 
This example demonstrates the ability of the Galilean approach to model temporal uncertainty in the same framework as uncertainty in position, velocity and orientation. 
The approach allows one to transform time-lagged measurements forward in time to be used in a current fusion algorithm (pre-integration), as well as providing a mechanism to model correlations between timestamp and spatial uncertainties in the fusion itself.  
These two examples are certainly not exhaustive, but they serve to demonstrate the potential benefits of exploiting Galilean symmetry and the matrix algebra of the Galilean group in solving robotics problems. 
We hope the paper serves as both an introduction to Galilean symmetry in robotics and a foundation for future work.

The remainder of the paper is organized as follows.
In Section \ref{sec:Galilean_transformations}, we give a brief introduction to the Galilean symmetry transformation and discuss \emph{events}, \emph{inertial velocities}, \emph{directions}, and event and velocity \emph{noise} as homogeneous coordinates for Galilean spacetime.
Next, in Section \ref{sec:Galilean_frames}, we consider Galilean frames as models for rigid-bodies moving in spacetime and show how the classical coordinate change matrix algebra used extensively in robotics generalises to Galilean transformation matrices. 
In Section \ref{sec:Kinematics}, we develop the kinematics of Galilean frames and write these kinematics in terms of the Lie-group algebra of the Galilean group. 
We apply this insight to the first example problem, modelling the motion of a rigid body with respect to an Earth-surface-fixed reference frame, in Section~\ref{sec:Rotating Earth}.
Finally, in Section \ref{sec:Fusion} we consider how the spacetime formulation of the Galilean group can be exploited to model noise in measurement timestamps.
We apply this to fusing measurements with uncertain timestamps with a state estimate that also has an uncertain timestamp.
A short conclusion follows in Section \ref{sec:Conclusion}. 

\section{Galilean Transformations}
\label{sec:Galilean_transformations}

We begin our exposition by defining Galilean transformations of space and time.
We assume that the reader is familiar with the classical special orthogonal $\SO(3)$ and special Euclidean $\SE(3)$ Lie groups used extensively in robotics but develop the Galilean-specific aspects largely from first principles in a tutorial style.
For more detailed background on Lie groups, we refer the interested reader to \cite{2005_Selig_Geometric,2021_Sola_Micro}.

\subsection{Inertial Reference Frames and Events}
Modelling of physical systems starts by defining coordinates with which to express physical quantities.
In classical robotics, a reference frame $\frameA$ is an origin along with a right-handed set of orthonormal directions that provide a coordinate system for space.
A point in space is given coordinates $\idx{p}{A}{}{}
= (x,y,z) \in \R^3$ that define its position with respect to the reference frame $\frameA$.
In Galilean spacetime one adds the concept of a time coordinate.
That is, a point $\idx{p}{A}{}{}$ occurs at time $\idx{t}{A}{}{}$ measured in seconds with respect to the time origin $t = 0$ of a frame $\frameA$.
Time in a different reference frame can be measured differently, although in Galilean relativity, all clocks run at the same rate and so it is only the time offset that can differ between frames.
The combination of a point $\idx{p}{A}{}{}$ in space occurring at time $\idx{t}{A}{}{}$ is termed an \emph{event} $\idx{p}{A}{}{t} = (\idx{p}{A}{}{} ,\idx{t}{A}{}{})$, all written with respect to a given reference frame $\frameA$.

\subsection{Inertial and Coordinate Velocity}

\begin{figure}
\centering
\input{figures/tikz/velocities_tikz_RAM.tex}
\caption{
Consider a point $p$ (black dot) with respect to a moving frame reference frame $\frameA$ (red axes).
The point is moving with linear velocity $v_p$ with respect to a third inertial observer. 
The frame is moving with linear velocity $w$ and rotating with angular velocity $\Omega$ with respect to the same inertial observer. 
The relative inertial velocity $v = (v_p - w)$ of the point is that induced only by velocity of the point and the translation of the frame. 
The coordinate velocity $\dot{p}$ of the point is the sum of the relative inertial velocity and the perceived coordinate velocity induced by rotation of the frame.}
\label{fig:velocity_of_point}
\end{figure}

Galilean spacetime differs from classical space in that it inherently involves motion.
The reference frame can be moving and points in space can be moving. 
Since only the relative motion between a moving point and the reference frame is observable, the motion of the reference frame is not modelled separately in Galilean symmetry, only the motion of points in space relative to a reference are considered. 
As long as the reference frame is moving with constant linear velocity and not rotating, then Newton's laws of motion hold in the relative coordinates. 
Such reference frames are termed \emph{inertial} reference frames and changing coordinates from one inertial reference frame to another may introduce a constant offset to the relative velocity of a point, but will not change the equations of motion. 
Indeed, such changes of coordinates are precisely the Galilean transformations introduced below.

In robotics, we will also need to consider coordinates that are not inertial. 
Let $\frameA$ be an arbitrary moving reference frame.
In Galilean spacetime, one defines the \emph{inertial velocity} $\idx{v}{A}{}{} = (v_x,v_y,v_z) \in \R^3$ of a point $\idx{p}{A}{}{}$ even if the reference frame used is not inertial. 
A key conceptual distinction is that the inertial velocity is not the same as the \emph{coordinate velocity} $\idx{\dot{p}}{A}{}{} \not= \idx{v}{A}{}{}$, except in the special case when the reference frame $\frameA$ is inertial.
Figure \ref{fig:velocity_of_point} illustrates the difference. 
From within the frame, it is impossible to perceive the motion of the frame itself and the only observable is the coordinate velocity $\dot{p}$. 
However, from a third person's perspective, anchored to an inertial frame, one can see the frame $\frameA$ translating and rotating, and it is possible to identify the relative inertial velocity $(v_p - w)$ of the point independently of the induced velocity $-\Omega \times p$ due to the frame rotation. 

To see that relative inertial velocity is \emph{intrinsic}, that is defined independently of the frames of reference, consider a fourth person's perspective, anchored to a second inertial frame, moving with constant velocity $u$ with respect to the third person's frame.   
The fourth person measures velocity $R v_p -u$ for the velocity of the point, and $R w - u$ for the velocity of the reference frame $\frameA$, where $R$ is the relative (constant) rotation between the inertial frames. 
However, the relative velocity 
\[
(R v_p - u) - ( Rw - u ) = R (v_p - w) = R v
\]
is the same physical quantity rotated into the fourth persons coordinates.
Note that if the frame $\frameA$ is inertial (with angular velocity $\Omega \equiv 0$) then the coordinate velocity of the point is $v = v_p - w$ is the relative inertial velocity of the point as expected. 
This discussion shows that there are two concepts for the relative velocity of a point in coordinates; the relative inertial velocity $(v_p - w)$ and the coordinate velocity $\dot{p}$. 
The (relative) inertial velocity is the intrinsic concept that leds to Galilean symmetry.

\subsection{Galilean Transformations} 
\label{sec:Galilean_Transformation}

\begin{figure}
\centering
\includegraphics[scale=0.55]{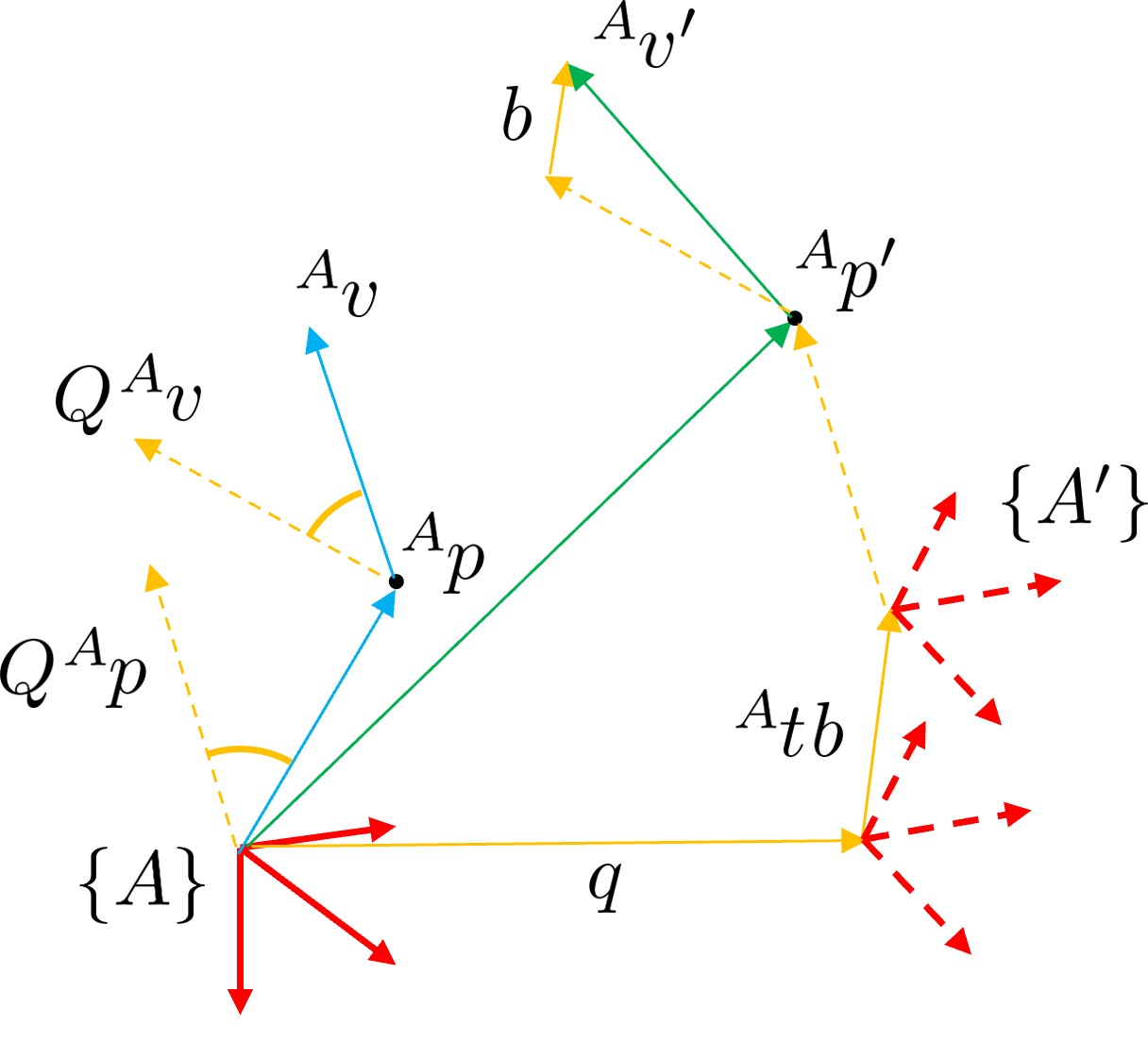}
\caption{
Galilean transformation $\idx{p}{A}{}{} \mapsto \idx[\prime]{p}{A}{}{}$ and $\idx{v}{A}{}{} \mapsto \idx[\prime]{v}{A}{}{}$. 
The point $\idx{p}{A}{}{}$ (blue) is rotated by $Q$ (dashed orange), translated by $q$ (orange), translated by $\idx{t}{A}{}{} b$ (orange) to become $\idx[\prime]{p}{A}{}{}$ (green).
The velocity  $\idx{v}{A}{}{}$ is rotated by $Q$ (dashed orange) then boosted by $b$ (orange) to become $\idx[\prime]{v}{A}{}{}$ (green). 
}
\label{fig:GalileanTransformation}
\end{figure}

A Galilean transformation of spacetime is a rotation $Q \in \SO(3)$, a spatial translation $q \in \grpR^3$, a temporal translation (or offset) $\delta \in \grpR$, and a velocity translation (or boost) $\bv \in \grpR^3$.
Here the bold face script denotes a group, the special orthogonal group $\SO(3)$ and the additive linear groups $\grpR$ and $\grpR^3$.
The rotation and spatial translation encode a classical rigid-body transformation of space.
The temporal translation changes the time at which the event occurs.
The boost models change in the inertial velocity of points in spacetime.
An event $\idx{p}{A}{}{t} = (\idx{p}{A}{}{}, \idx{t}{A}{}{}) \in \R^4$ and its inertial velocity $\idx{v}{A}{}{}$ are transformed by $(Q, \bv, q, \delta)$ according to (see Figure~\ref{fig:GalileanTransformation}) 
\begin{subequations}\label{eq:Galilean_transformation}
\begin{align}
\idx[\prime]{p}{A}{}{} & := Q \idx{p}{A}{}{} + q + \idx{t}{A}{}{} \bv    \label{eq:Galilean_transformation_xi}\\
\idx[\prime]{t}{A}{}{} & :=
\idx{t}{A}{}{} + \delta  \label{eq:Galilean_transformation_t}\\
\idx[\prime]{v}{A}{}{} & :=
Q \idx{v}{A}{}{} + \bv
\label{eq:Galilean_transformation_v}
\end{align}
\end{subequations}
Unpacking \eqref{eq:Galilean_transformation_xi} one sees that the coordinates of the point are transformed by a rigid-body transformation $Q \idx{p}{A}{}{} + q$ plus the term $\idx{t}{A}{}{} \bv $.
Since Galilean transformations model constant linear motion relative to the origin of the reference frame, the boost $\bv$ needs to be considered over the time history of the motion in the reference frame, that is from $t=0$. 
Specifically, in the transformed coordinates, the point $p$ has already been moving with additional velocity $\bv$ since the time origin $t=0$ of frame $\frameA$, generating the offset $\idx{t}{A}{}{} \bv $ (Fig.~\ref{fig:GalileanTransformation}).
The time offset \eqref{eq:Galilean_transformation_t} is the easiest to understand and simply shifts when the event occurs in the frame $\frameA$ time coordinates.
The transformed velocity is also straightforward: the old velocity is rotated and the additional  inertial velocity due to the boost is added.

\subsection{Homogeneous Coordinates and the Galilean Group}
\label{sec:homogeneous_Matrices}

Similar to rigid-body transformations, there is a realisation of the Galilean transformation in homogeneous coordinates.
We write homogeneous coordinates $\idx{\ob{p}}{A}{}{t} = (x,y,z,t,1) \in \R^5$ for the event $\idx{p}{A}{}{t} = (\idx{p}{A}{}{},\idx{t}{A}{}{})$.
The final coordinate of an event in homogeneous spacetime coordinates is always unity and plays a similar role to the $(x,y,z,1) \in \R^4$ homogeneous coordinates used in classical robotics texts.
The inertial velocity associated with an event also has homogeneous coordinates $\idx{\ob{v}}{A}{}{} := (v, 1, 0) \in \R^5$.
Here we place a zero in the last entry, indicating a velocity, and a one in the 4th entry, that can be conceptualised as $\dot{t} = 1$.

A Galilean transformation matrix is defined to be
\begin{align}
G = \begin{pmatrix}
Q & \bv & q \\ 0 & 1 & \delta \\ 0 & 0 & 1
\end{pmatrix} \in \R^{5 \times 5}. 
\label{eq:homogeneous_matrix}
\end{align}
The inverse 
\begin{align}
G^{-1} =
\begin{pmatrix}
Q^{\top} & -Q^{\top}\bv & -Q^{\top}\left(q - \delta \bv \right) \\
0 & 1 & - \delta \\
0 & 0 & 1
\end{pmatrix}.
\label{eq:homogeneous_matrix_inverse}
\end{align}
is also a Galilean transformation matrix. 
It is easily verified that if $G_1, G_2$ satisfies \eqref{eq:homogeneous_matrix} then $G_1 G_2$ satisfies \eqref{eq:homogeneous_matrix} and the set of Galilean transformation matrices 
\begin{align}
\Gal(3) = \cset{G \in \R^{5 \times 5}}{G \text{ satisfies } \protect\eqref{eq:homogeneous_matrix}}
\label{eq:Galilean_group}
\end{align}
with inverse \eqref{eq:homogeneous_matrix_inverse} and the matrix identity $I_5$ is a matrix Lie-group. 

A Galilean transformation matrix encodes the Galilean spacetime transformations \eqref{eq:Galilean_transformation} through linear matrix multiplication
\begin{subequations}
\label{eq:linear_action}
\begin{align}
\begin{pmatrix}
Q & \bv & q \\ 0 & 1 & \delta \\ 0 & 0 & 1
\end{pmatrix}
\begin{pmatrix}
p \\t \\ 1
\end{pmatrix}
& =
\begin{pmatrix}
Q p + t \bv + q \\ t + \delta \\ 1
\end{pmatrix},
\label{eq:linear_action_event}\\[1mm]
\begin{pmatrix}
Q & \bv & q \\ 0 & 1 & \delta \\ 0 & 0 & 1
\end{pmatrix}
\begin{pmatrix}
v \\ 1 \\ 0
\end{pmatrix}
& =
\begin{pmatrix}
Q v + \bv  \\ 1 \\ 0
\end{pmatrix}.
\label{eq:linear_action_velocity}
\end{align}
\end{subequations}
It follows that $\Gal(3)$ is a matrix realisation of the group of Galilean transformations of spacetime.
In particular, we will identify the matrix Lie-group $\Gal(3)$ \eqref{eq:Galilean_group} as the Galilean group of spacetime transformations, analogous to the way $\SE(3)$ is identified with the group of rigid-body pose transformations in classical robotics. 

Directions $\idx{\eta}{A}{}{} = (\eta_x,\eta_y,\eta_z) \in \Sph^2$ are also important and can be used to model measurements and states such as magnetometer readings, star sights, the direction of gravity, etc.
Directions transform purely by rotation $\idx{\eta}{A}{}{} \mapsto
Q \idx{\eta}{A}{}{}$.
The homogeneous coordinates of a direction are $\idx{\ob{\eta}}{A}{}{}
= (\idx{\eta}{A}{}{},0,0)$ and the homogeneous transform matrix also transforms directions
\begin{align}
\begin{pmatrix}
Q & \bv & q \\ 0 & 1 & \delta \\ 0 & 0 & 1
\end{pmatrix}
\begin{pmatrix}
\eta \\ 0 \\ 0
\end{pmatrix}
& =
\begin{pmatrix}
Q \eta   \\ 0 \\ 0
\end{pmatrix}.
\label{eq:linear_action_direction}
\end{align}

In summary, homogeneous coordinates for Galilean spacetime have events $\ob{p}_t = (p,t,1)$, inertial velocities $\ob{v} = (v,1,0)$, directions $\ob{\eta} = (\eta, 0 , 0)$. 
All of these classes of objects transform according to the Galilean spacetime transformations \eqref{eq:Galilean_transformation} under linear multiplication by the Galilean transformation matrix and the set of all such transformation matrices is known as the Galilean group. 

\subsection{Spatio-Temporal Uncertainty}
\label{sec:Noise}

The final category of homogeneous coordinates that we consider is associated with \emph{event noise} and \emph{velocity noise}. 
A measurement of position $p$ at time $t$ with Gaussian noise in both the position and timestamp can be modeled by an uncertain event $y_t = p_t + \mu \in \R^4$ where $\mu \sim \GP(0, \Sigma)$ for some covariance $\Sigma \in \R^{4 \times 4}$ is event noise. 
Let $\mu \in \R^4$ be event noise then its homogeneous coordinates are written 
\begin{align} 
\ob{\mu} = \begin{pmatrix} \mu \\ 0 \end{pmatrix} \in \R^5, 
\label{eq:homog_free_event}
\end{align}
where the fifth entry is zero. 

\begin{figure}
\centering
\includegraphics[scale=0.55]{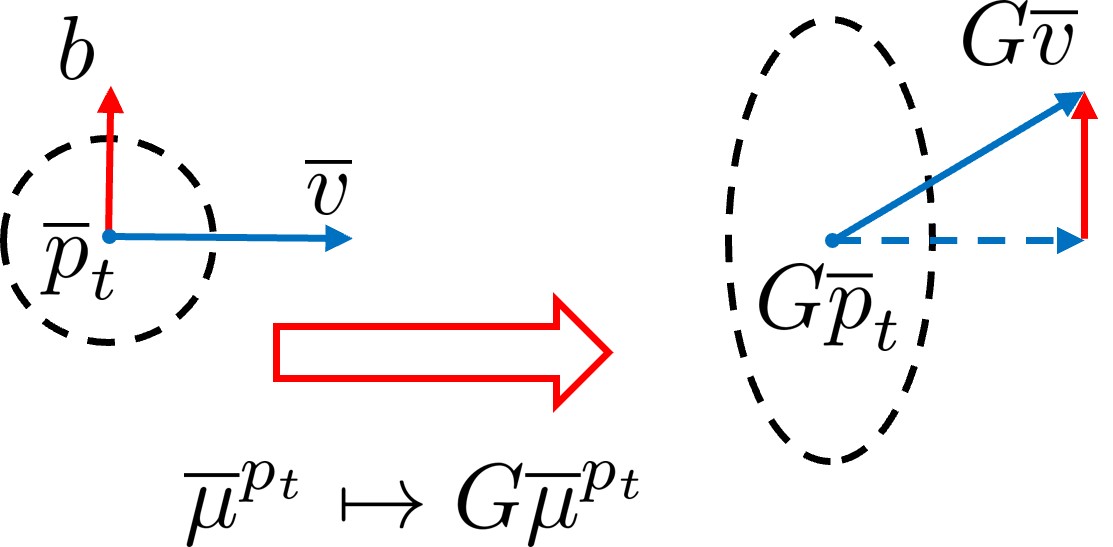}
\caption{
Spatio-temporal coupling of timestamp and velocity uncertainty under Galilean transformation.
Homogeneous uncertainty $\ob{\mu}^{p_t}$ in the event $\ob{p}_t$ (shown by the dashed covariance ellipse on the left) is smeared by the Galilean transformation $G$ in the boost direction (shown by the expanded dashed covariance ellipse on the right).  
The figure also shows transformation of the velocity $\ob{v}$ to $G \ob{v}$ with the addition of the boost. 
}
\label{fig:SmearedNoise}
\end{figure}

Homogeneous coordinates of event noise also transform under multiplication by a Galilean transformation matrix.  
That is, for a Galilean transformation matrix $G$ then $G \ob{y}_t = G \ob{p}_t + G \ob{\mu}^{p_t}$. 
In particular, for $\mu^{p_t} = (\mu^p, \mu^t) \in \R^4$ with $\mu^p \in \R^3$ and $\mu^t \in \R$, then 
\begin{align}
G \ob{\mu} = 
\begin{pmatrix}
Q & \bv & q \\ 0 & 1 & \delta \\ 0 & 0 & 1 
\end{pmatrix}
\begin{pmatrix}
\mu^p \\ \mu^t \\ 0  
\end{pmatrix}
= 
\begin{pmatrix}
Q \mu^p + \bv \mu^t \\ \mu^t \\ 0  
\end{pmatrix}
\label{eq:uncertainty_smearing}
\end{align}
and the temporal uncertainty $\mu^t$ is smeared by the velocity boost $\bv$ to create additional uncertainty in the position variable; see Figure~\ref{fig:SmearedNoise}. 

Noise in velocity $\mu^v \in \R^3$ is modelled by \emph{velocity noise}.
The homogeneous coordinates for velocity noise $\ob{\mu}^v = (\mu^v,0 , 0)$ have two zeros in the fourth and fifth positions.
Velocity noise also transforms under multiplication by Galilean transformation matrix acting on the homogeneous coordinates, although in this case the transformation is just multiplication by the rotation of the Galilean transformation, the same as simply modelling the change of basis of the velocity representation in classical robotics texts \cite{barfoot2024state}. 
Noise in directions $\eta \in \Sph^2 \hookrightarrow \R^3$ cannot be modelled directly as an additive term in the homogeneous coordinates due to the manifold constraint of the sphere (cf.~\S\ref{sec:Frame-uncertainty}). 

\subsection{Galilean Lie Algebra}
\label{sec:Lie_Algebra}

The Lie algebra of the matrix Lie group $\Gal(3)$ can be modeled as the matrix subspace corresponding to the tangent space $\tT_I \Gal(3)$ at the identity element,
\begin{align}
\gal(3) = \cset{U \in \R^{5 \times 5}}{U = 
\begin{pmatrix}
\omega^\wedge & a & w \\ 
0 & 0 & \kappa \\ 
0 & 0 & 0   
\end{pmatrix}
},
\label{eq:U_gothgal}
\end{align}
where $a, w \in \R^3$, $\kappa \in \R$, and 
\begin{align}
\omega^\wedge := \begin{pmatrix}
0 & - \idx{\omega}{}{}{3} &   \idx{\omega}{}{}{2} \\
\idx{\omega}{}{}{3} & 0 & - \idx{\omega}{}{}{1}  \\
- \idx{\omega}{}{}{2} & \idx{\omega}{}{}{1} & 0 
\end{pmatrix},
\label{eq:wedge_skew}
\end{align}
with $\omega = (\idx{\omega}{}{}{1}, \idx{\omega}{}{}{2}, \idx{\omega}{}{}{3})$.

The algebra models infinitesimal motion or displacement on the Lie group. 
For kinematic inputs, then $\omega$ is an angular velocity and $a$ is an acceleration.
In certain applications the $w$ term can be used to encode an offset velocity (such as a wind or current), and $\kappa$ could be used to encode clock drift.
The Lie-algebra is also used to encode spatio-temporal uncertainty in events, as discussed in Section \S\ref{sec:Noise}, and uncertainty in Galilean frames themselves, which will be covered in Section \S\ref{sec:Frame-uncertainty}. 
For noise, all four terms $(\omega,a,w,\kappa)$ are important and encode error in attitude, velocity, position and time. 

The `wedge' operator $\left(\cdot\right)^\wedge : \R^{10} \rightarrow \gal(3)$ is defined to be the linear map 
\begin{align}
(\omega,a,w,\kappa)^\wedge = U = \begin{pmatrix}
\omega^\wedge & a & w \\ 
0 & 0 & \kappa \\ 
0 & 0 & 0   
\end{pmatrix}\in \gal(3).
\label{eq:wedge_operator}
\end{align}
The inverse `vee' operator $(\cdot)^\vee : \gal(3) \to \R^{10}$ is defined by $U^\vee = (\omega,a,w,\kappa) \in \R^{10}$, where $U$ is as written in \eqref{eq:wedge_operator}.

\section{Galilean Frames}
\label{sec:Galilean_frames}

Motion of a rigid-body in spacetime can be modelled by attaching a Galilean frame rigidly to the object. 
A Galilean frame consists of an event representing the origin of the frame in spacetime, the relative inertial velocity of the origin point, and a right-handed set of three orthonormal directions defining a coordinate system.
A rigid-body at time $t$ is modelled as spacetime trajectory such that the origin of its rigidly attached frame passes through an event $(p, t)$, with instantaneous velocity $v$, and has orientation $R$, at time $t$. 
The construction is entirely analogous to representing the pose of a rigid-body by attaching a frame (an origin and a right-handed set of orthonormal directions) used in classical robotics except that a Galilean frame representation adds velocity and time. 

Coordinates for a Galilean frame, orientation $R$, velocity $v$, position $p$ and time $t$, can only be written as relative coordinates. 
That is, the coordinate expressions for $(R,v,p,t)$ must be written with respect to some other Galilean frame that we term the reference frame. 
Let $\frameA$ denote the reference frame and $\frameB$ denote the target frame of interest.
The relative position and time of the target frame are written as an event in spacetime coordinates $(\idx{p}{A}{}{B},\idx{t}{A}{}{B})$ relative to the origin (at zero time) of the reference frame.
The coordinates of the orthonormal basis 
\[
\idx{R}{A}{}{B} = (\idx{\eb}{A}{B}{1}, \idx{\eb}{A}{B}{2}, \idx{\eb}{A}{B}{3}) \in \SO(3),
\]
are the coordinate expressions of axes of $\frameB$ expressed with respect to frame $\frameA$. 
The inertial velocity $\idx{v}{A}{}{B}$ is also expressed with respect to $\frameA$. 

The same matrix structure \eqref{eq:homogeneous_matrix_inverse} used to encode a Galilean transformation is ideal for encoding a Galilean frame; 
\[
\idx{F}{A}{}{B} =
\begin{pmatrix}
\idx{R}{A}{}{B} & \idx{v}{A}{}{B} &  \idx{p}{A}{}{B} \\
0 & 1 &  \idx{t}{A}{}{B} \\
0 & 0 & 1
\end{pmatrix} \in \R^{5 \times 5} 
\]
where the first three columns are homogeneous directions (columns of $\idx{R}{A}{}{B}$), the fourth column is a homogeneous inertial velocity $\idx{v}{A}{}{B}$, and the final column is a homogeneous event $(\idx{p}{A}{}{B},\idx{t}{A}{}{B})$. 
The approach is entirely analogous to using homogeneous transformation matrices to provide coordinates of the pose of a rigid-body in classical robotics. 

\begin{remark} 
[Frame versus Pose] 
Formally, a Galilean frame is a set of coordinates in spacetime while a Galilean pose is associated with a rigid-body moving in space time. 
Since the pose of the rigid-body is coordinatised by attaching a Galilean frame rigidly to the body then the mathematical representation of the two concepts is identical.
\QED
\end{remark} 

A frame can be transformed by a Galilean transformation $G \in \Gal(3)$ (cf.~Fig.~\ref{fig:GalileanTransformation}) 
\begin{align*}
G \idx{F}{A}{}{B} & =
\begin{pmatrix}
Q & \bv & q \\ 0 & 1 & \delta \\ 0 & 0 & 1
\end{pmatrix}
\begin{pmatrix}
\idx{R}{A}{}{B} & \idx{v}{A}{}{B} & \idx{p}{A}{}{B} \\ 0 & 1 & \idx{t}{A}{}{B} \\ 0 & 0 & 1
\end{pmatrix} \\[1mm]
& =
\begin{pmatrix}
Q \idx{R}{A}{}{B} & Q \idx{v}{A}{}{B} + \bv & Q \idx{p}{A}{}{B} + \idx{t}{A}{}{B} \bv + \idx{p}{A}{}{B} \\ 0 & 1 & \idx{t}{A}{}{B} + \delta \\ 0 & 0 & 1
\end{pmatrix}.
\end{align*}
Note that the origin transforms as \eqref{eq:linear_action_event}, the velocity transforms according to \eqref{eq:linear_action_velocity}, while the directions encoded in the columns of $\idx{R}{A}{}{B}$ transform as directions \eqref{eq:linear_action_direction}.

\begin{remark}[Group elements versus frames]\label{rem:groupvsframe}
Using homogeneous coordinates to represent Galilean frames introduces a conceptual challenge: Galilean group elements and Galilean frames are both written in homogeneous matrix coordinates and are easily confused with each other.
This difficulty is also present in classical robotics where rigid-body transformations and pose are both written as homogeneous matrices. 
The distinction is extremely important and in the following text we will be careful to distinguish between a \emph{Galilean transformation} $G = (Q,\bv ,q,\delta) \in \Gal(3)$ and a \emph{Galilean frame} $F = (R,v,p,t)$. 

To add to the confusion, and in common with the conventions for pose $P \in \SE(3)$ in classical robotics, we will write $F \in \Gal(3)$ for the coordinates of a Galilean frame to indicate that they have the structure of a Galilean homogeneous matrix, even though they are not an element of the group. 
Formally, Galilean frames are elements of the Galilean torsor, that is, the manifold structure of $\Gal(3)$ without the group structure. 
\QED
\end{remark} 

\subsection{Change of Coordinates}
\label{sec:Coordinate_Change}

Analogous to rigid-body pose coordinates written as transformation matrices, Galilean frames have an interpretation as coordinate mappings.
Consider an event $\idx{\ob{p}}{B}{}{t} = (\idx{p}{B}{}{},\idx{t}{B}{}{},1)$ indicating a point $\idx{p}{B}{}{t}$ occurring at $\idx{t}{B}{}{}$ seconds with respect to a  Galilean frame $\frameB$. 
Then 
\begin{align*}
\idx{F}{A}{}{B} \idx{\ob{p}}{B}{}{t}
& = 
\begin{pmatrix}
\idx{R}{A}{}{B} & \idx{v}{A}{}{B} &  \idx{p}{A}{}{B} \\
0 & 1 &  \idx{t}{A}{}{B} \\
0 & 0 & 1
\end{pmatrix}
\begin{pmatrix}
\idx{p}{B}{}{} \\ \idx{t}{B}{}{}  \\ 1 
\end{pmatrix} \\
&= 
\begin{pmatrix}
\idx{R}{A}{}{B}  \idx{p}{B}{}{} +  \idx{p}{A}{}{B} + \idx{t}{B}{}{} \idx{v}{A}{}{B}  \\  
\idx{t}{A}{}{B} + \idx{t}{B}{}{}\\ 1 
\end{pmatrix}
= 
\begin{pmatrix}
\idx{p}{A}{}{t} \\  \idx{t}{A}{}{B}\\ 1 
\end{pmatrix} 
= 
\idx{\ob{p}}{A}{}{t} 
\end{align*}
and the event expressed in the reference frame has the temporal state $\idx{t}{A}{}{B}
+ \idx{t}{B}{}{}$ as expected. 
In general, it is easily verified that the homogeneous coordinates 
\begin{align*}
\idx{\ob{p}}{A}{}{t} & = \idx{F}{A}{}{B} \idx{\ob{p}}{B}{}{t}
\end{align*}
for any event $\idx{\ob{p}}{B}{}{t}$ written in $\frameB$ homogeneous coordinates.
One also gets
\begin{align*}
\idx{\ob{v}}{A}{}{} & = \idx{F}{A}{}{B} \idx{\ob{v}}{B}{}{} \quad \text{~and} \quad 
\idx{\ob{\eta}}{A}{}{} = \idx{F}{A}{}{B} \idx{\ob{\eta}}{B}{}{}
\end{align*}
for changing coordinates of inertial velocities and directions, 
and analogous mappings for event and velocity noise. 

Note that these coordinate mappings are not Galilean spacetime transformations. 
Specifically, the Galilean frame $\idx{F}{A}{}{B}$ can be applied to change coordinate representation of an event, velocity or direction, to express \emph{the same physical variable} in different coordinates; while a Galilean transformation $G \in \Gal(3)$ is frame agnostic, and transforms events, velocities and directions \emph{to new physical variables} --- events, velocities and directions.
This distinction is analogous to that of using homogeneous pose matrices as coordinate maps versus rigid-body transformations of space (screw theory in classical robotics), and it is one of the reasons why it is so important to distinguish between Galilean transformations $G$ and Galilean frames $F$. 

Concatenating and inverting these coordinate mappings, and exploiting the matrix group algebra of the Galilean group, leads to the frame transformation equations 
\begin{subequations}
\label{eq:frame_transformations}
\begin{align}
\idx[-1]{F}{A}{}{B} & = \idx{F}{B}{}{A}  \label{eq:frame_transformation_inverse} \\
\idx{F}{A}{}{C} & = \idx{F}{A}{}{B} \idx{F}{B}{}{C} \label{eq:frame_transformation_product}
\end{align}
\end{subequations}
that are analogous to the well known frame algebra used extensively in rigid-body modelling in robotics. 

The frame index notation is very powerful and incredibly useful in many situations.
It is also cumbersome and obscures the simplicity of many of the equations that follow.
Where it is obvious from context which frames are being considering, we will often drop the indices, typically writing $F = \idx{F}{A}{}{B}$ where the indices $\frameA$ and $\frameB$ are clear from context. 
However, we will never confuse the notational distinction between a Galilean transformation $G \in \Gal(3)$ and a Galilean frame $F$.

\subsection{Galilean versus Inertial Frames}
\label{sec:Inertial_Frames}

A Galilean frame is like an event in that it exists only at a specific time. 
Given two Galilean frames $\frameA$ and $\frameB$, the relative coordinates $\idx{F}{A}{}{B}$ encode the fixed time offset $\idx{t}{A}{}{B}$ of $\frameB$ with respect to frame $\frameA$. 
Any Galilean frame, however, corresponds to a \emph{unique} inertial motion. 

Let $\frameB$ be a Galilean frame defined with respect to $\frameA$, that is, with coordinates $\idx{F}{A}{}{B}$. 
There is a unique inertial frame $\{C_t\}$ defined (with respect to $\frameA$) for all time $t \in \R$ defined by Galilean coordinates 
\begin{align}
\idx{F}{A}{}{C_t}(t) = \begin{pmatrix} \idx{R}{A}{}{B} & \idx{v}{A}{}{B} & \idx{p}{A}{}{B} + t \idx{v}{A}{}{B} \\
    0 & 1 & \idx{t}{A}{}{B} + t \\ 
    0 & 0 & 1 
\end{pmatrix}. 
\label{eq:unique_inertial}
\end{align}
The frame $\{C_t\}$ defined by these coordinates is an inertial frame, that is it is non-rotating and moves with constant linear velocity. 
Note that $\{C_0\} = \frameB$ at time $t = 0$.

Let $\frameZero$ be a Galilean reference frame and consider the relative coordinates $\idx{F}{0}{}{B}$ that encode the coordinates of $\frameB$ with respect to frame $\frameZero$.
The Galilean reference $\frameZero$ corresponds to an unique inertial frame $\{A_t\}$, colocated with $\frameZero = \{A_0\}$ at zero time and not moving or rotating \eqref{eq:unique_inertial}.  
The coordinates of this inertial frame with respect to $\frameZero$ are 
\[
\idx{F}{0}{}{A_t} = \begin{pmatrix} \mathbf{I}_3 & 0 & 0 \\ 0 & 1 & t \\ 0 & 0 & 1 \end{pmatrix}. 
\]  
where $\mathbf{I}_3$ is the $3\times3$ identity matrix. 
If $\idx{\ob{p}}{A_t}{}{t} = (p,0,0)$ are the coordinates of an event at the current time in the inertial frame $\{A_t\}$; that is, the event occurs at the present time $t$ in $\{A_t\}$ and so its time coordinate in homogeneous coordinates is zero. 
Then the coordinates of this event in the Galilean reference frame $\frameZero$ are 
\[
\idx{\ob{p}}{0}{}{t} = \idx{F}{0}{}{A_t} \idx{\ob{p}}{A_t}{}{t} = (p,t,0) 
\]
as expected. 
That is, the coordinates of the point $\idx{p}{A_t}{}{} = \idx{p}{0}{}{}$ and it is only the time index that distinguishes the inertial frame coordinates from the reference frame $\frameZero$. 
Analogous relationships hold for velocities (without the time index) and for rotations. 
It follows that intuition about inertial motion, that is, motion relative to an inertial frame $\{A_t\}$, written in coordinates, corresponds in a natural manner to modelling kinematics of motion in the Galilean reference $\frameZero = \{A_0\}$. 
This insight is used in Section \S\ref{sec:Kinematics}. 

\subsection{Galilean Frame Uncertainty}
\label{sec:Frame-uncertainty}

A key advantage of representing robot states using Galilean frames rather than classical $\SE(3)$ pose is the ability to model spatio-temporal uncertainty. 
Specifically, it is possible to build information state models that incorporate dependence between the spatial states orientation and position, with the kinematic state of velocity, and the temporal state of time. 

\begin{figure}
\centering
\includegraphics[scale=0.55]{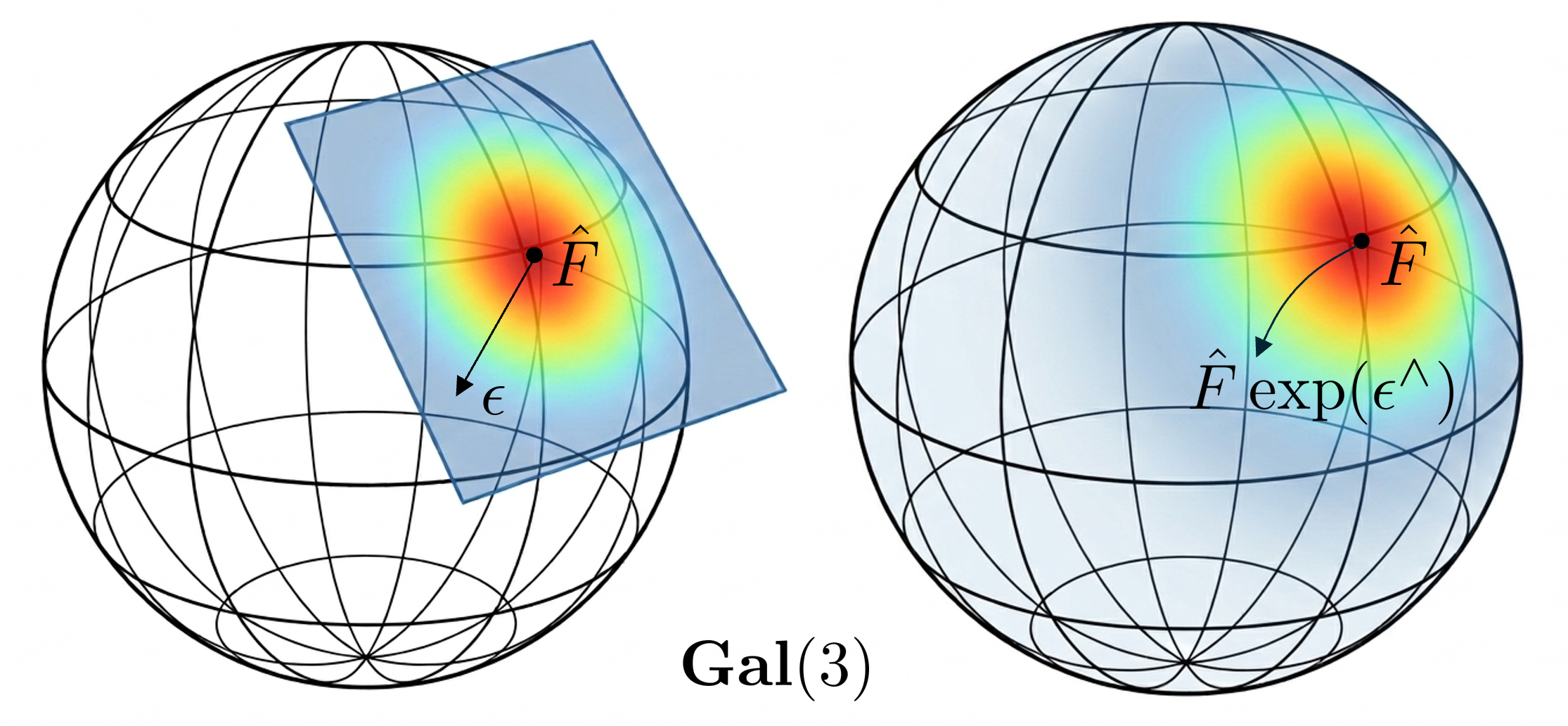}
\caption{
Representation of a concentrated Gaussian on $\Gal(3)$. 
The Gaussian distribution is defined in Euclidean coordinates $\epsilon \in \R^n$ on the tangent space $\tT_{\hat{F}} \grpG$ of the Lie-group and then mapped onto a distribution on the group itself using group multiplication, the wedge map \eqref{eq:wedge_operator}, and the exponential map. 
The Lie-group $\Gal(3)$ is drawn as a sphere for simplicity of the diagram and the reader should be aware that the geometry of $\Gal(3)$ is more complex than shown here. 
}
\label{fig:ConcentratedGaussian}
\end{figure}

We use the structure of a concentrated Gaussian \cite{Wang_2006_Error,2024_Ge_LCSS} (see Figure~\ref{fig:ConcentratedGaussian}) to encode uncertainty in a Galilean frame 
\begin{align}
F = \hat{F}\exp(\epsilon^\wedge), \quad \epsilon \sim \GP(0, Q),
\label{eq:F_RandVar}
\end{align}
where $\hat{F}=  (\hat{R}, \hat{v}, \hat{p}, \hat{t}) \in \Gal(3)$ is the state estimate parameter, $Q \in \R^{10 \times 10}$ is a positive definite covariance matrix, and $\epsilon$ is a Gaussian random variable. 
The exponential\footnote{
In general, the expression for the matrix exponential is defined by an infinite matrix series.
However, in common with $\SO(3)$ and $\SE(3)$, there are simplifications of the general formulae to algebraic combinations of classical (transcendental) trigonometric functions for $\Gal(3)$. 
A number of authors have recently developed such expressions for $\exp$, $\log$ and various left and right Jacobians on $\Gal(3)$ 
\cite{barfoot2025integral,Delama_2024,2024_Kelly_arxiv,2021_Giefer_Uncertainties}. 
For the matrix exponential, we recommend \cite[Eq.~8]{Delama_2024}. 
}
$\exp$ is the classical matrix exponential on the Galilean group representation.

The random variable $\epsilon \in \mathbb{R}^{10}$ is written in Euclidean coordinates but corresponds to group coordinates through 
\eqref{eq:F_RandVar}. 
The process is analogous to building a Gaussian model in the group's tangent space and then draping it, via the algebra wedge representation and exponential map, over the manifold at $\hat{F}$ (Figure~\ref{fig:ConcentratedGaussian}). 
As long as the Gaussian density is \emph{concentrated} around the origin of $\gothgal(3)$ then any overlap where the exponential map is not uniquely defined (e.g., due to extremely large rotational or other uncertainties) is not stochastically significant. 
This stochastic model is a powerful tool in engineering applications due to the simplicity of the representation of covariance \eqref{eq:F_RandVar}, and is preferable to alternatives such as the Fisher, Bingham, \cite{kent1982fisher} or von Mises \cite {maksimov1967necessary} distributions, or their more general Langevin distributions \cite{watson1984theory,chirikjian2009stochastic}. 

\begin{figure}
\centering
\includegraphics[scale=0.55]{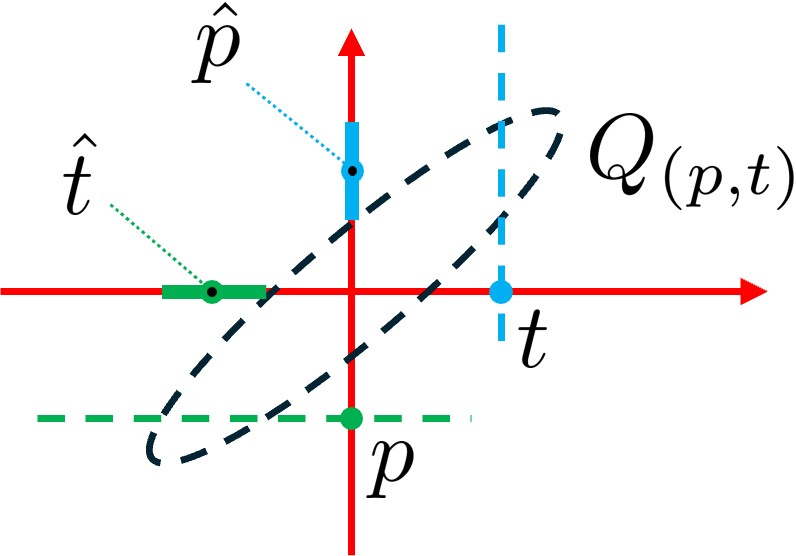}
\caption{
Spatio-Temporal stochastic correlation.  
Here the $(7\!:\!9,10)$ and $(10,7\!:\!9)$ entries of $Q \in \R^{10 \times 10}$ are non-zero and lead to the covariance correlation $Q_{(p,t)}$ shown between the position $p$ and the time $t$ variables. 
If time $t$ is measured (blue point on the time-axis) then the uncertainty in position $\hat{p}$ is the conditional distribution shown in the heavy blue line  (indicating a collapsed ellipse)  on the $p$-axis. 
If position $p$ is known (green point on the position-axis) then the uncertainty in time $\hat{t}$ is the conditional distribution shown in the heavy green line on the time axis. 
} 
\label{fig:StochasticCorrelation}
\end{figure}

Uncertainty in the timestamp $t$ is encoded in the $\kappa$ entry of the $\epsilon^\wedge$ algebra element. 
Importantly, this uncertainty can be correlated with uncertainty in the navigation states $R$, $v$, and $p$, of the vehicle through off diagonal elements in $Q$, specifically the tenth row and column of $Q$. 
As a consequence, measurement of time can correlate to information about the navigation states $R$, $v$ and $p$, or vice versa. 
Figure~\ref{fig:StochasticCorrelation} indicates how stochastic correlation between time and navigation states can be exploited for state estimation. 
Spatio-temporal correlation of uncertainty is common in timed positioning systems such as Global Navigation Satellite Systems or robotic sensors with time delays (cf.~\S\ref{sec:Fusion}).

\section{Galilean Kinematics}
\label{sec:Kinematics}

Frame kinematics is a fundamental tool used to model motion in robotics.
The structure of kinematics represented using the matrix Lie-group structure of the Galilean group is simple and highly intuitive.
Indeed, the elegance and power of the kinematics of motion expressed in the Galilean formulation is one of our main motivations in writing this paper. 

\subsection{Kinematics of Galilean Frames}
\label{sec:Kinematics_Origin}

The Galilean frame coordinates of a moving frame $\frameB = \{B_t\}$ with respect to reference Galilean frame $\frameZero$ are written 
\[
F = \idx{F}{0}{}{B} = \begin{pmatrix}
R & v & p \\ 0 & 1 & t \\ 0 & 0 & 1
\end{pmatrix},
\]
where $R$, $v$ and $p$ are the relative orientation, inertial velocity, and position, and $t$ is the relative time associated with the Galilean frame $\frameB$. 
We drop indices in the variables $F = (R,v,p,t)$ to simplify the following notation. 

Consider the unique inertial frame $\frameA = \{A_t\}$, colocated with the reference $\frameZero$ at time zero (cf.~\S\ref{sec:Inertial_Frames}). 
The coordinates $R$, $v$ and $p$ are defined with respect to the reference $\frameZero$, but correspond to the same physical variables defined with respect to the inertial frame $\frameA$. 
The expressions for the pose kinematics with respect to an inertial frame are well known \cite{2002_Goldstein_mechanics}.
We add time kinematics $\dot{t} = 1$ since all clocks run at the same time in the Galilean universe to obtain 
\begin{align}
\dot{R} = R \omega^\wedge, \quad \dot{v} = R a, \quad \dot{p} = v, \quad \dot{t} = 1, 
\label{eq:kinematics_coordinate}
\end{align}
where the wedge operator $\wedge$ was defined earlier \eqref{eq:wedge_skew}. 
Here, the \emph{Galilean angular velocity} $\omega = \idx{\omega}{B}{A}{B} \in \R^3$ and the \emph{Galilean acceleration} $a = \idx{a}{B}{A}{B}$ encode the relative change in motion of the moving frame $\frameB$ (bottom right index), with respect to the \emph{inertial} frame $\frameA$ (bottom left index), expressed in the body coordinates $\frameB$ (top left index). 
The angular velocity $\omega$ and acceleration $a$ capture the non-inertial motion, that is,the deviation from an inertial motion, of the Galilean frame at time $t$. 
The linear kinematics $\dot{p} = v$ are equivalent to the coordinate kinematics since inertial frame constructed is non-rotating (Fig.~\ref{fig:velocity_of_point} with $\Omega \equiv 0$). 

\begin{figure}
\centering
\includegraphics[scale=0.55]{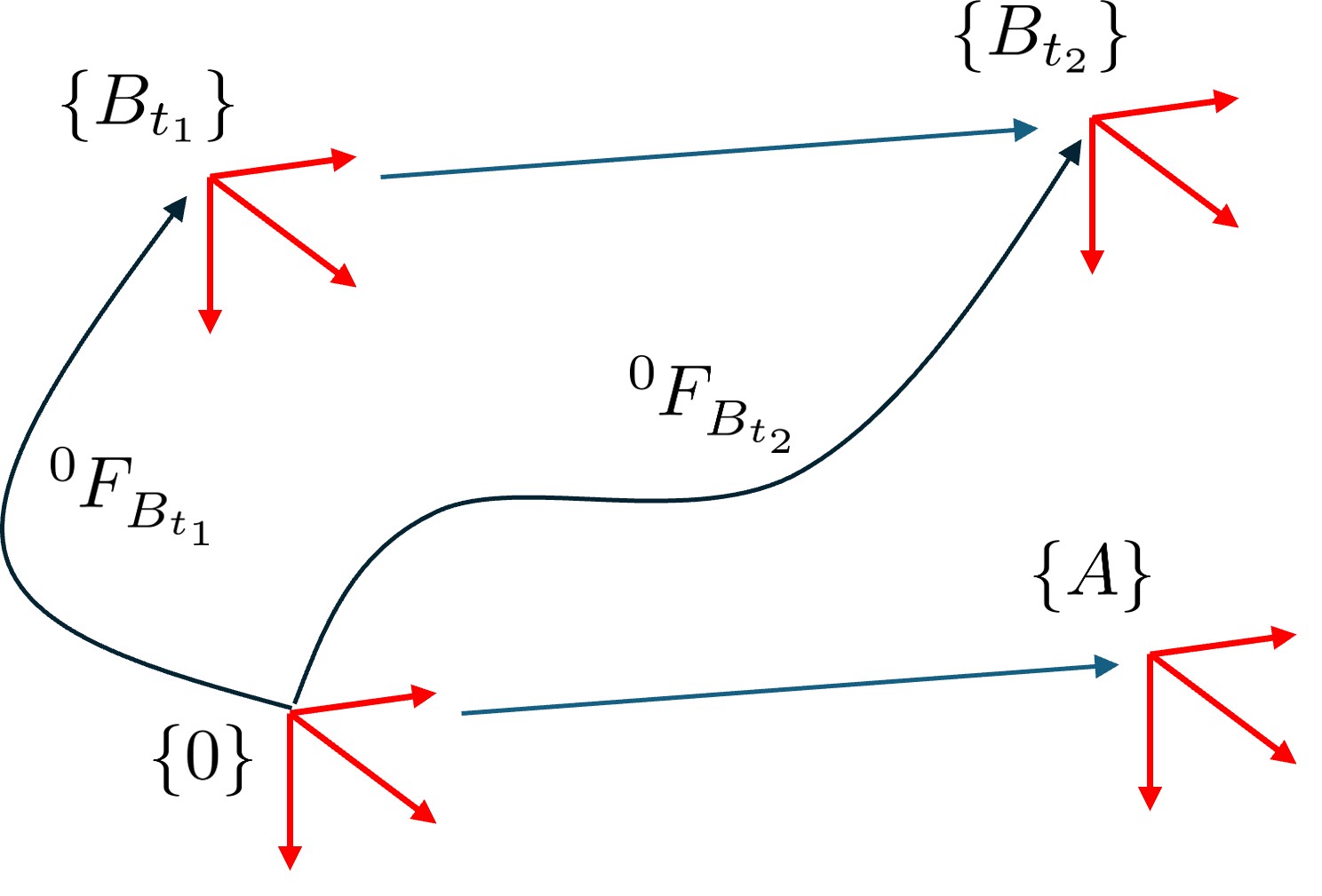}
\caption{
Kinematics of a Galilean Frame.   
The reference $\frameZero$ is fixed in time, while frame $\frameA$ is the associated stationary inertial frame. 
The Galilean frame $\frameB$ is moving in space-time.
The kinematics of the Galilean frame coordinates $F = \idx{F}{0}{}{B_t}$ with respect to the reference are given by \eqref{eq:kinematics_Galilean_matrix}. 
} 
\label{fig:GalileanTransformation_zero_ref}
\end{figure}

The kinematics \eqref{eq:kinematics_coordinate} can be written directly in the homogeneous coordinates $F = \idx{F}{0}{}{B}$.
Computing $\ddt F$ as a matrix equation, one obtains
\begin{align}
\begin{pmatrix}
\dot{R} & \dot{v} & \dot{p} \\
0 & 0 & \dot{t} \\
0 & 0 & 0
\end{pmatrix}
& =
\begin{pmatrix}
R\omega^\wedge & R a & v \\
0 & 0 & 1 \\
0 & 0 & 0
\end{pmatrix} \\[1mm]
& =
\begin{pmatrix}
R & v & p \\
0 & 1 & t \\
0 & 0 & 1
\end{pmatrix}
\begin{pmatrix}
\omega^\wedge &  a & 0 \\
0 & 0 & 1 \\
0 & 0 & 0
\end{pmatrix}.
\label{eq:kinematics_Galilean}
\end{align}
Define the input matrices
\begin{align}
U(\omega, a)  & = \idx{U}{B}{A}{B} =
\begin{pmatrix}
\omega^\wedge & a & 0 \\
0 & 0 & 0 \\
0 & 0 & 0
\end{pmatrix}
\in \gal(3)  \label{eq:V}, \\[1mm]
N & =  \begin{pmatrix}
0 & 0 & 0 \\
0 & 0 & 1 \\
0 & 0 & 0
\end{pmatrix}
\in \gal(3),
\label{eq:N}
\end{align}
where $U = U(\omega,a)$ captures the dependence on the accelerations and angular velocities and $N$ captures the time kinematics $\dot{t} = 1$ as well as modelling the linear kinematics $\dot{p} = v$.
The resulting Galilean kinematics (with respect to a reference frame) can be written simply 
(cf.~Fig.~\ref{fig:GalileanTransformation_zero_ref})
\begin{align}
\dot{F} = F\left(U(\omega, a) + N\right).
\label{eq:kinematics_Galilean_matrix}
\end{align} 
This equation has the structure of a left-invariant system on a matrix Lie-group leading to a range of nice properties. 

\begin{remark} 
Although we write $U(\omega,a)$ and $N$ as elements in $\gal(3)$, they are not actually elements of the Lie-algebra since $F$ is a Galilean frame and not a Galilean transformation (Remark~\ref{rem:groupvsframe}). 
The kinematics \eqref{eq:kinematics_Galilean} exploits the algebra of the homogeneous matrix Lie-group realisation of $\Gal(3)$, and it follows that the inputs $U(\omega,a)$ and $N$ have the structure corresponding to the matrix realisation of the $\gal(3)$.
We will use the notation $\gal(3)$ for both inputs to frame kinematics and Lie-algebra elements since the distinction does not merit additional notation.
\QED
\end{remark} 

Note that the Galilean angular velocity $\omega_B$ and Galilean acceleration $a_B$ are intrinsic quantities. 
To see this, consider two inertial reference frames $\{A_1\}$ and $\{A_2\}$ associated with two reference frames $\{0_1\}$ and $\{0_2\}$. 
Then, since $\{A_1\}$ and $\{A_2\}$ are inertial, there is a constant Galilean transformation $G$ such that 
\[
\idx{F}{A_2}{}{B} = G\, \idx{F}{A_1}{}{B}
\]
for all time, and hence $\idx{F}{0_2}{}{B} = G\, \idx{F}{0_1}{}{B}$. 
Taking the time differential, one has 
\begin{align}
\idx{\dot{F}}{0_2}{}{B} & = G\, \idx{\dot{F}}{0_1}{}{B} \notag  \\
\idx{F}{0_2}{}{B} U(\idx{\omega}{B}{A_2}{B},\idx{a}{B}{A_2}{B}) 
& = G \, \idx{F}{0_1}{}{B}U(\idx{\omega}{B}{A_1}{B},\idx{a}{B}{A_1}{B}) \notag \\
\idx{F}{0_2}{}{B} U(\idx{\omega}{B}{A_2}{B},\idx{a}{B}{A_2}{B}) 
& = \idx{F}{0_2}{}{B}U(\idx{\omega}{B}{A_1}{B},\idx{a}{B}{A_1}{B}) \notag \\
U(\idx{\omega}{B}{A_2}{B},\idx{a}{B}{A_2}{B}) 
& = U(\idx{\omega}{B}{A_1}{B},\idx{a}{B}{A_1}{B}) \label{eq:intrinsic_Galilean_velocity}
\end{align}
since $\idx{F}{0_2}{}{B}$ is invertible. 
It follows that $\omega_B := \idx{\omega}{B}{A_2}{B} = \idx{\omega}{B}{A_1}{B}$ and 
$a_B :=  \idx{a}{B}{A_2}{B} = \idx{a}{B}{A_1}{B}$ do not depend on the choice of inertial reference frame.
Since Galilean inputs are intrinsic, we define the Galilean input matrix 
\begin{align} 
U_B := U(\omega_B,a_B) = U(\idx{\omega}{B}{A}{B},\idx{a}{B}{A}{B})
\label{eq:U_B_def}
\end{align}
where $\frameA$ is any inertial frame.

\subsection{Isochronous Kinematics of Galilean Frames: Inertial Reference Case}
\label{sec:Kinematics_Inertial}

We say that two frames are \emph{isochronous} if they have the same time-index. 
The relative coordinates of a Galilean frame $\frameB$ with respect to an isochronous reference $\frameA$ are 
\begin{align}
\idx{F}{A}{}{B} & = \begin{pmatrix}
\idx{R}{A}{}{B} & \idx{v}{A}{}{B} & \idx{p}{A}{}{B} \\
0 & 1 & 0 \\
0 & 0 & 1
\end{pmatrix}. 
\label{eq:SE23_pose}
\end{align}
In particular, the relative time-offset is zero since the isochronous reference and the Galilean frame have the same time index. 
The set of all such matrices is a sub-group of the Galilean group that is known as the \emph{isochronous Galilean group} \cite{levy1971galilei}.
The group was recently identified by Barrau \etal \cite{barrau2014invariant,barrau2016invariant,Barrau_2023}, termed the \emph{double direct spatial isometry} group, as playing a key role in the development of high-performance inertial navigation systems.
It is most commonly referred to as the \emph{extended pose} group \cite{2020_Barrau_Mathematical,brossard2021associating,FornasierEquivariantSymmetries2025} in the robotics literature. 
We use the notation of Barrau \etal \cite{barrau2014invariant,barrau2016invariant,Barrau_2023} 
\begin{align}
\SE_2(3) = \cset{
  \begin{pmatrix}
Q & \bv & q \\ 
0 & 1 & 0 \\
0 & 0 & 1  
\end{pmatrix}}
{Q \in \SO(3), \bv \in \grpR^3, q \in \grpR^3 }.
\label{eq:extendedpose}
\end{align}

\begin{figure}
\centering
\includegraphics[scale=0.55]{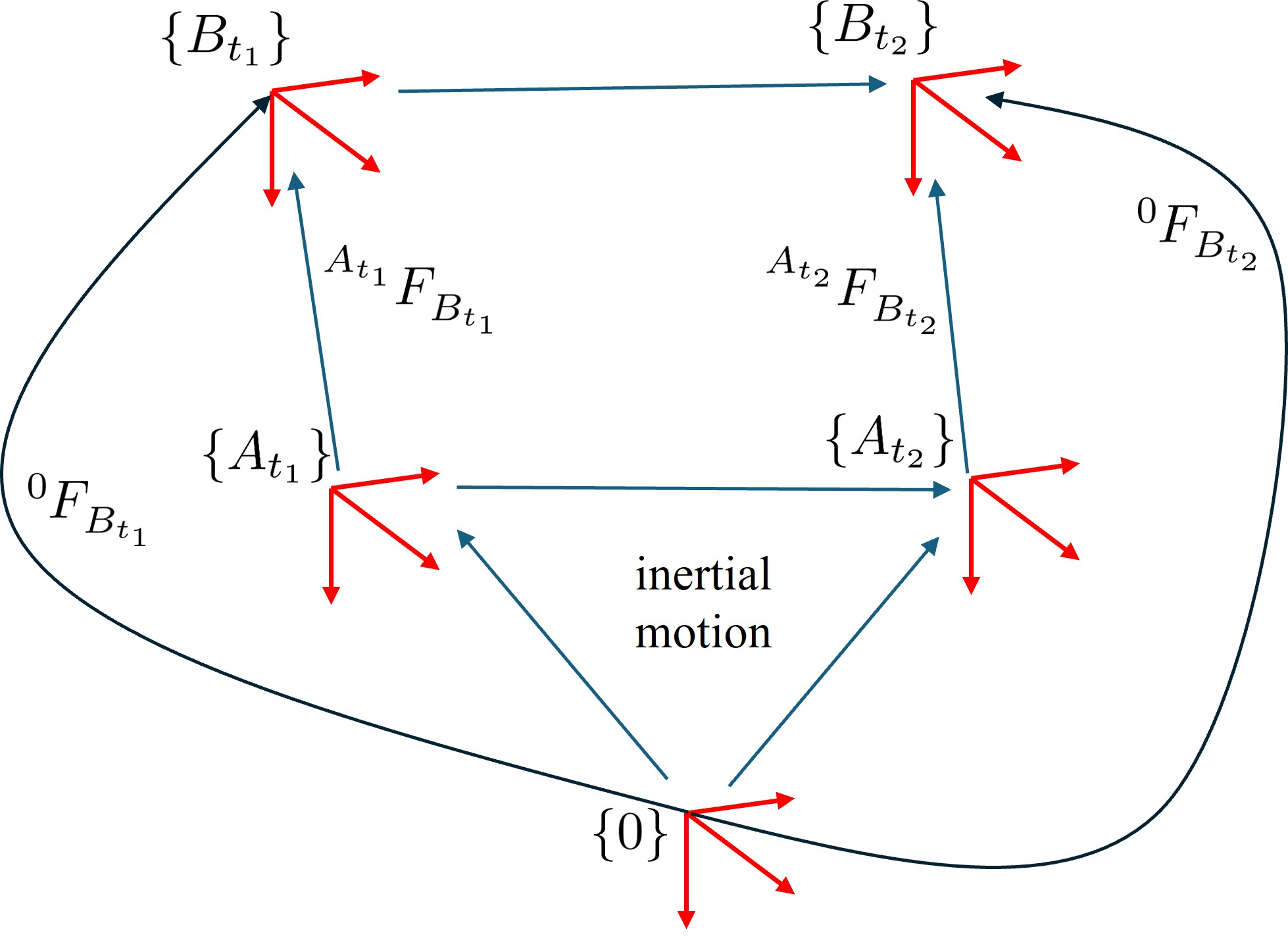}
\caption{
Isochronous Inertial Kinematics of a Galilean Frame.   
An inertial reference $\frameA$ moves with respect to the origin of $\frameZero$. 
The moving target Galilean frame $\frameB$ is isochronous with respect to the inertial frame $\frameA$. 
The kinematics of the isochronous Galilean frame $\idx{F}{A_t}{}{B_t}$ is given by \eqref{eq:isochronous_inertial_Galilean_kinematics}. 
} 
\label{fig:GalileanTransformation_inertial_ref}
\end{figure}

Consider a Galilean reference $\frameZero$ fixed in time and express $\frameA$ and $\frameB$ with respect to $\frameZero$; 
\[
\idx{F}{0}{}{A}  = \begin{pmatrix}
\idx{R}{0}{}{A} & \idx{v}{0}{}{A} & \idx{p}{0}{}{A} \\
0 & 1 & \idx{t}{0}{}{A} \\
0 & 0 & 1
\end{pmatrix}, 
\quad 
\idx{F}{0}{}{B}  = \begin{pmatrix}
\idx{R}{0}{}{B} & \idx{v}{0}{}{B} & \idx{p}{0}{}{B} \\
0 & 1 & \idx{t}{0}{}{B} \\
0 & 0 & 1
\end{pmatrix}
\]
where $\idx{t}{0}{}{A} = \idx{t}{0}{}{B}$ since $\frameA$ and $\frameB$ are isochronous.

Assume frame $\frameA$ is inertial.
Recalling \eqref{eq:kinematics_Galilean_matrix}, the kinematics with respect to the Galilean frame $\frameZero$ for frame $\idx{F}{0}{}{A}$ and $\idx{F}{0}{}{B}$ are given by 
\begin{subequations}
\label{eq:zero_reference_kinematics}
  \begin{align}
\idx{\dot{F}}{0}{}{A} & =\idx{F}{0}{}{A} N  \label{eq:zero_reference_kinematics_A} \\
\idx{\dot{F}}{0}{}{B} & = \idx{F}{0}{}{B} \left( U(\omega_B,a_B)  + N\right) 
\label{eq:zero_reference_kinematics_B}
\end{align}
\end{subequations}
where $N$ is given by \eqref{eq:N}, $(\omega_B, a_B)$ are the Galilean inputs for $\frameB$. 
Note that the Galilean angular velocity and acceleration of the inertial frame $\frameA$ are zero since it is inertial.  
From here, computing the kinematics of $\idx{F}{A}{}{B} = \idx[-1]{F}{0}{}{A} \idx{F}{0}{}{B}$ and recalling $\ddt(X^{-1}) = -X^{-1}{\dot X}X^{-1}$ one has (cf.~Fig.~\ref{fig:GalileanTransformation_inertial_ref})
\begin{align}
\ddt \idx{F}{A}{}{B}
& = \idx[-1]{\dot{F}}{0}{}{A}\idx{F}{0}{}{B} + \idx[-1]{F}{0}{}{A}\idx{\dot{F}}{0}{}{B} \notag \\
& = - \idx[-1]{F}{0}{}{A}
\idx{\dot{F}}{0}{}{A}\idx[-1]{F}{0}{}{A} \idx{F}{0}{}{B}
+ \idx{F}{0}{}{A}\idx{F}{0}{}{B} (U(\omega_B,a_B)+ N)  \notag\\
& = - \idx[-1]{F}{0}{}{A}
\idx{F}{0}{}{A}  N \idx{F}{A}{}{0}\idx{F}{0}{}{B} \notag\\ 
& \makebox[2cm]{}+ \idx{F}{A}{}{B} U(\omega_B,a_B) + \idx{F}{A}{}{B}  N  \notag \\
& = \idx{F}{A}{}{B} N - N \idx{F}{A}{}{B} + \idx{F}{A}{}{B} U(\omega_B,a_B) 
\label{eq:derive_isochronous_inertial_Galilean_kinematics}
\end{align}
Simplifying this by writing $F = \idx{F}{A}{}{B}$ one obtains the \emph{isochronous Galilean frame kinematics} with respect to an inertial reference 
\begin{align}
\dot{F} = FN - N F + F U(\omega_B,a_B)
\label{eq:isochronous_inertial_Galilean_kinematics}
\end{align}

Since this is a tutorial paper there is value in writing out these computations in detail to see how the classical coordinate equations are recovered. 
Firstly, the linear kinematics are generated by the term 
\begin{align}
F N - N F & =
\begin{pmatrix}
R & v & p \\
0 & 1 & 0 \\
0 & 0 & 1
\end{pmatrix}
\begin{pmatrix}
0 & 0 & 0 \\ 0 & 0 & 1 \\ 0 & 0 & 0
\end{pmatrix} \\
& \makebox[2cm]{}
-
\begin{pmatrix}
R & v & p \\
0 & 1 & 0 \\
0 & 0 & 1
\end{pmatrix}
\begin{pmatrix}
0 & 0 & 0 \\ 0 & 0 & 1 \\ 0 & 0 & 0
\end{pmatrix} \notag \\
&
= \begin{pmatrix}
0 & 0 & v \\
0 & 0 & 1 \\
0 & 0 & 0
\end{pmatrix}
-
\begin{pmatrix}
0 & 0 & 0 \\
0 & 0 & 1 \\
0 & 0 & 0
\end{pmatrix} \notag \\
& =
\begin{pmatrix}
0 & 0 & v \\
0 & 0 & 0 \\
0 & 0 & 0
\end{pmatrix}.
\label{eq:FN-NF_expression}
\end{align}
That is, this term encodes the relative motion $\dot{p} = v$ between the two frames $\frameA$ and $\frameB$.
The Galilean angular velocity $\omega_B$ and Galilean acceleration $a_B$ 
lead to 
\begin{align*}
F U(\omega_B, a_B) & =
\begin{pmatrix}
R & v & p \\
0 & 1 & 0 \\
0 & 0 & 1
\end{pmatrix}
\begin{pmatrix}
(\omega_B)^\wedge & a_B & 0 \\
0 & 0 & 0 \\
0 & 0 & 0
\end{pmatrix} \\[1mm]
& =
\begin{pmatrix}
R (\omega_B)^\wedge & R a_B & 0 \\
0 & 0 & 0 \\
0 & 0 & 0
\end{pmatrix}
\end{align*}
encodes the body-fixed exogenous motion of the active agent.
That is \eqref{eq:isochronous_inertial_Galilean_kinematics} encodes the classical Galilean kinematics \eqref{eq:kinematics_coordinate} \cite{2002_Goldstein_mechanics}. 

\begin{remark} 
It is interesting to note that $U, N \in \gal(3)$ but although $U \in \gothse_2(3)$, the time matrix $N \not\in \gothse_2(3)$: the (4,5) element of $N$, $\kappa$ in \eqref{eq:U_gothgal}, should be zero for $\se_2(3)$ \cite{barrau2016invariant,Barrau_2023}. 
This is due to the fact that the isochronous Galilean group does not model the time coordinate. 
The matrix kinematics \eqref{eq:zero_reference_kinematics_A} still hold, since the combined term $FN - N F$ is a vector field on $\SE_2(3)$.
Indeed, this term constructs the linear kinematics $\dot{p} = v$ \eqref{eq:FN-NF_expression}.
For those interested in the underlying Lie theory of this structure, details can be found in van Goor \etal \cite{van2021autonomous}.
\QED
\end{remark}

\subsection{Isochronous Kinematics of Galilean Frames: Moving Reference Case}
\label{sec:Kinematics_NonInertial}

Once the kinematics of Galilean frames with respect to an isochronous inertial reference are established it is possible to extend this model to Galilean frames moving with respect to non-inertial reference frames.  
This includes reference frames that are rotating and accelerating and situations where there is gravity. 

\begin{figure}
\centering
\includegraphics[scale=0.55]{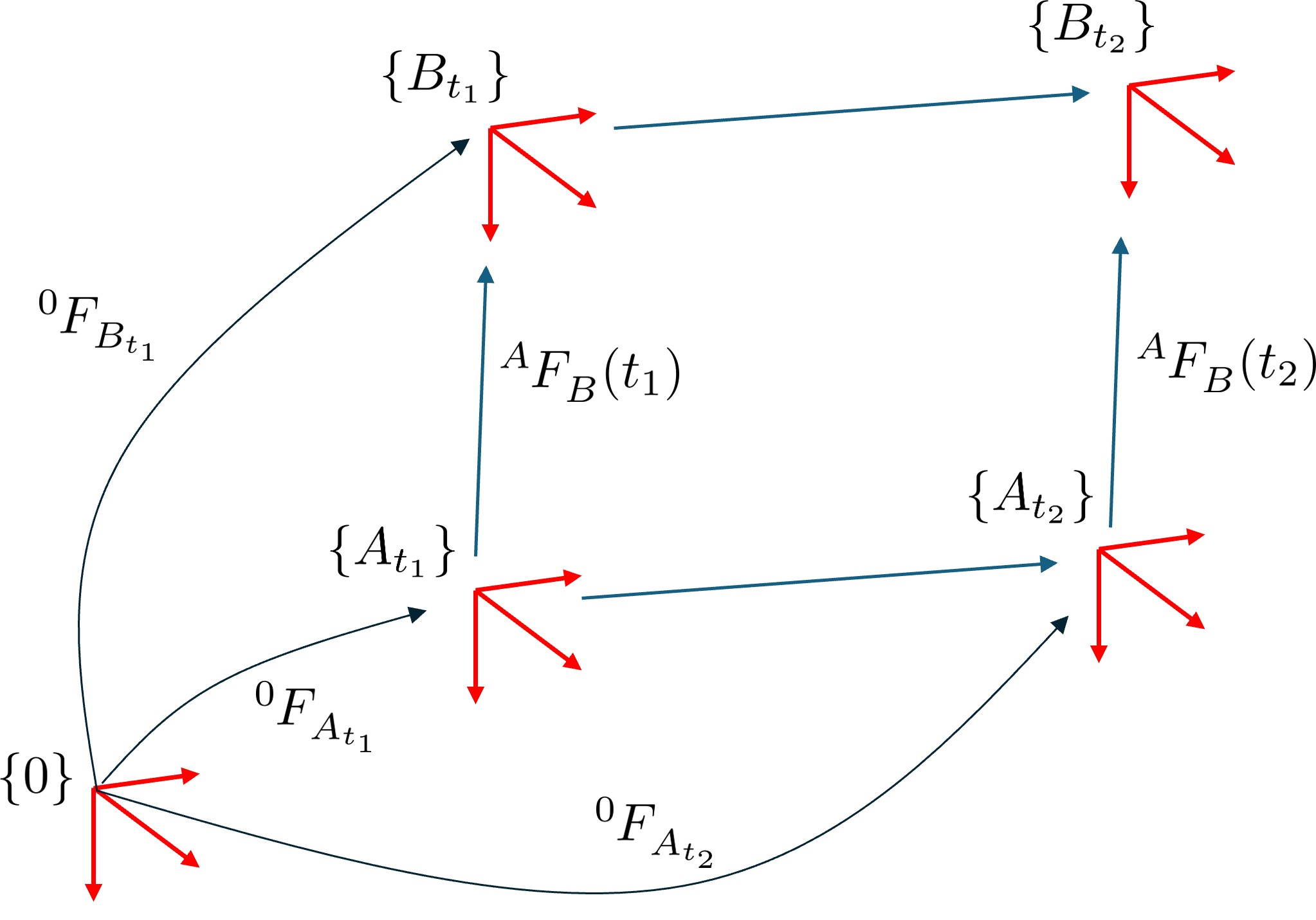}
\caption{
Isochronous Galilean kinematics.   
Both the reference $\frameA$ and target $\frameB$ Galilean frames are moving with respect to $\frameZero$. 
The kinematics of the isochronous Galilean frame $\idx{F}{A_t}{}{B_t}$ is given by \eqref{eq:IsochronousKinematics}. 
}
\label{fig:Isochronous_Kinematics}
\end{figure}

Consider a non-inertial reference frame $\frameA$. 
The kinematics of a general moving reference frame $\frameA$ with respect to a fixed Galilean frame $\frameZero$ are 
\begin{align}
\idx{\dot{F}}{0}{}{A} & =\idx{F}{0}{}{A} (U(\omega_A, a_A) + N)   
\label{eq:zero_reference_kinematics_A2} 
\end{align}
where $\omega_A$ and $a_A$ are the Galilean angular velocity and acceleration respectively of $\frameA$ (cf.~\eqref{eq:zero_reference_kinematics_B}). 
Using the same derivation \eqref{eq:derive_isochronous_inertial_Galilean_kinematics} it is easily verified that the kinematics of the isochronous Galilean frame $F = \idx{F}{A}{}{B}$ with respect to a non-inertial reference $\frameA$ are given by (cf.~Fig.~\ref{fig:Isochronous_Kinematics})
\begin{align}
\dot{F} = F N - N F - U(\omega_A,a_A) F + F U(\omega_B, a_B). 
\label{eq:IsochronousKinematics}
\end{align}

It is interesting to consider what equations \eqref{eq:IsochronousKinematics} 
look like in component form rather than matrix form.
The new term compared to \eqref{eq:isochronous_inertial_Galilean_kinematics} has the form 
\begin{align*}
- U(\omega_A,a_A) F
& =
- \begin{pmatrix}
(\omega_A)^\wedge & a_A & 0 \\
0 & 0 & 0 \\
0 & 0 & 0
\end{pmatrix}
\begin{pmatrix}
R & v & p \\
0 & 1 & 0 \\
0 & 0 & 1
\end{pmatrix} \\
& =
\begin{pmatrix}
-(\omega_A)^\wedge R & - \omega_A \times v - a_A& -\omega_A \times p \\
0 & 1 & 0 \\
0 & 0 & 1
\end{pmatrix}.
\end{align*}
since $\omega^\wedge a = \omega \times a$ for any vectors $\omega$ and $a$.
Collecting the terms for each of the elements of 
\eqref{eq:IsochronousKinematics} one obtains 
\begin{subequations}\label{eq:rotating_earth_kinematics_components}
\begin{align}
\ddt {R} & = - (\omega_A)^\wedge R + R (\omega_B)^\wedge  \label{eq:rotating_earth_kinematics_components_R}\\
\ddt{v} & =  - \omega_A \times  v - a_A + R a_B \label{eq:rotating_earth_kinematics_components_v}\\
\ddt{p} & = - \omega_A \times p + v. \label{eq:rotating_earth_kinematics_components_p}
\end{align}
\end{subequations}
A key observation is that \eqref{eq:rotating_earth_kinematics_components_p} correctly models the coordinate velocity $\dot{p}$ from the inertial velocity $v$ (see Fig.~\ref{fig:velocity_of_point}). 
The simplicity of this formulation in capturing the relative coordinate velocity while modelling inertial velocity is major advantage of this formulation. 

\begin{remark}
Setting the Galilean angular velocity and acceleration of the reference frame to zero, $\omega_A = 0 = a_A$, in \eqref{eq:rotating_earth_kinematics_components}, one recovers the inertial kinematics \eqref{eq:isochronous_inertial_Galilean_kinematics} (see also \eqref{eq:kinematics_coordinate}) as one would expect. 
\end{remark}

It is important to remember that the Galilean accelerations $a_A$ and $a_B$ must include any gravitational specific forces that are experienced by frames $\frameA$ and $\frameB$, noting that an inertial frame cannot experience gravitational force by definition. 
Let $g_A$ and $g_B$ denote the gravitational specific forces on frame $\frameA$ and $\frameB$ expressed in their respective body-frames.
The \emph{proper acceleration}. or ``the acceleration relative to free fall'', is defined to be 
\[
a_A^\text{p}  := a_A - g_A \quad \text{ and } \quad 
a_B^\text{p}  := a_B - g_B 
\]
for frames $\frameA$ and $\frameB$ respectively. 
Separating out the proper acceleration and gravitational specific force in \eqref{eq:IsochronousKinematics} one obtains 
\begin{align}
\dot{F} = & F N - N F - U(\omega_A,a_A^\text{p}) F + F U(\omega_B, a_B^\text{p}) \notag \\ 
& \qquad\qquad - \left( U(0,g_A) F -  F U(0, g_B) \right). 
\label{eq:isochronous_kinematics_2}
\end{align}

It is informative to look at the gravitation specific force terms together. 
Note that since $F$ is isochronous then 
\begin{align*}
F U(0,g_B) = & 
\begin{pmatrix}
R & v & p \\
0 & 1 & 0 \\
0 & 0 & 1
\end{pmatrix}
\begin{pmatrix}
0 &  g_B & 0 \\
0 & 0 & 0 \\
0 & 0 & 0
\end{pmatrix} \\
= & 
\begin{pmatrix}
0 &  R g_B & 0 \\
0 & 0 & 0 \\
0 & 0 & 0
\end{pmatrix}. 
\end{align*} 
while 
\begin{align*}
U(0,g_A) F  = & 
\begin{pmatrix}
0 &  g_A & 0 \\
0 & 0 & 0 \\
0 & 0 & 0
\end{pmatrix}
\begin{pmatrix}
R & v & p \\
0 & 1 & 0 \\
0 & 0 & 1
\end{pmatrix} \\ 
= & 
\begin{pmatrix}
0 &  g_A & 0 \\
0 & 0 & 0 \\
0 & 0 & 0
\end{pmatrix}. 
\end{align*} 
That is 
\[
 U(0,g_A) F -  F U(0, g_B) = 
 \begin{pmatrix}
0 &  g_A - R g_B & 0 \\
0 & 0 & 0 \\
0 & 0 & 0
\end{pmatrix}
\]
Consider a situation where two vehicles, modelled by frames $\frameA$ and $\frameB$, are moving in a constant gravitational field, a common approximation made when studying relative motion in robotics. 
Then one has $\idx{R}{A}{}{B} \idx{g}{}{}{B} = \idx{g}{}{}{A}$ since the gravitational specific force is the same for both vehicles and the terms 
$\idx{g}{}{}{B}$ and $\idx{g}{}{}{A}$ are expressed in body-coordinates. 
Recalling \eqref{eq:isochronous_kinematics_2} one obtains 
\begin{align}
\dot{F} & = FN - N F - U(\omega_A,a_A^\text{p})F + F U(\omega_B,\idx[\text{p}]{a}{}{}{B}). 
\label{eq:IsochronousKinematics_3}
\end{align}
In particular, if the same gravitational field is experienced by both frames, the gravitational field does not enter into the relative Galilean kinematics (of the extended pose) at all. 
Intuitively, if both vehicles are in free fall and isochronous, then their relative motion is unaffected by gravity. 
It is only when one of the frames is held is position against the force of gravity, such as an Earth fixed reference frame, or when the frames are far enough apart that the gravitational field no longer cancels, such as the case for satellites or global navigation problems, that the gravitational terms need be considered. 
Note that this relationship also fails if $F = \idx{F}{A}{}{B}$ includes a time-offset (cf.~\S\ref{sec:PreIntegration}). 

\section{Modelling Kinematics in a Rotating Earth Frame}
\label{sec:Rotating Earth}

To understand the power of modelling using the Galilean perspective we consider the classical problem of modelling the kinematics of a moving frame $\frameB$ attached to robotic vehicle with respect to an Earth fixed frame $\frameA$ at a known location. 
The goal of the model is to write down the kinematics of $\idx{F}{A}{}{B} \in \SE_2(3)$ as a function of IMU measurements and known parameters such as the rotation and gravitational constant of the Earth. 
Although this problem is fundamental in the domain of field robotics and the design inertial navigation systems, modelling the effect of centripetal and gravitation terms is complex and often poorly understood. 
The Galilean framework provides a simple and intuitive way to to understand and model these effects. 

Consider a reference frame $\frameA$ that is isochronous with the moving frame $\frameB$; that is fixed to, and rotates with, the surface of the Earth (Figure~\ref{fig:rotating_Earth}). 
We use the aerospace convention and consider the $z$-axis of the reference frame to be normal to the Earths surface and pointing inwards. 
To leverage the full Galilean framework we introduce an arbitrary third Galilean reference frame $\frameZero$ to write Galilean coordinates $\idx{F}{0}{}{A}$ and $\idx{F}{0}{}{B}$ for frames $\frameA$ and $\frameB$. 
However, note that the relative Galilean frame $\idx{F}{A}{}{B} = \idx[-1]{F}{0}{}{A}\idx{F}{0}{}{B}$ is defined independently of $\frameZero$ and the time offset of $\idx{F}{A}{}{B}$ is zero. 

\begin{figure}
\centering
\input{figures/tikz/kinematics_tikz_RAM.tex}
\caption{Galilean kinematics of an agent with respect to a moving frame attached to the surface of the Earth.}
\label{fig:rotating_Earth}
\end{figure}

The key steps in modelling motion of a vehicle in a rotating Earth frame for the purposes of inertial navigation is to understand the measurements provided by an inertial measurement unit and how to model gravity, contact, and centripetal forces for the Earth fixed frame. 

\subsection{Inertial Measurement Units}
\label{sec:IMU}

The Galilean angular velocity $\omega_B = \idx{\omega}{B}{0}{B}$ of a Galilean frame $\frameB$, 
expressed in the moving frame $\frameB$, is the rate of change of the relative orientation of the frame with respect to an \emph{inertial} reference $\frameZero$. 
An inertial measurement unit exploits the underlying physics of the intrinsic nature \eqref{eq:intrinsic_Galilean_velocity} of the Galilean angular velocity to measure  $\omega_B$ in the body frame $\frameB$ coordinates.
Indeed, a rate-gyroscope such as is used on a standard strap-down inertial measurement unit directly measures Galilean angular velocity 
\[
\omega^\text{IMU}_B = \omega_B + \text{noise}. 
\]

The Galilean acceleration $a_B = \idx{a}{B}{0}{B}$ of $\frameB$, expressed in the moving frame $\frameB$, is the rate of change of the \emph{inertial} velocity of the frame, 
defined with respect to an \emph{inertial} reference $\frameZero$. 
As shown earlier \eqref{eq:intrinsic_Galilean_velocity} this value is also intrinsic. 
However, due to the nature of gravitational force there is no inertial sensor technology that allows $a_B$ to be measured. 
The Galilean acceleration decomposes into two terms 
\[
a_B = a_B^\text{p} + g_B 
\]
the \emph{proper acceleration}  $a^\text{p}_B$, often referred to as ``the acceleration relative to free fall'', and the \emph{gravitational acceleration} $g_B$.
The proper acceleration can be measured by a mass-spring damper accelerometer and is available as an output from a six degree-of-freedom inertial measurement unit 
\[
a^\text{IMU}_B = a^\text{p}_B + \text{noise}. 
\]
The gravitational acceleration $g_B$ cannot be measured directly, and must be modelled. 
For robotics applications where an agent has an insignificant mass and is moving in the gravitational field due to the mass distribution of the Earth or other solar bodies whose positions are assumed to be known, the gravitational acceleration can be modelled as a function $g_B = g_B(p) \in \R^3$ of position, where we assume that $g_B$ is written in the body-coordinates of frame $\frameB$. 
Close to the Earths surface one can use a model such as an official Earth Gravitational Model EGM2008 \cite{pavlis2012development}. 
Recalling \eqref{eq:U_B_def} then 
\begin{align}
U(\omega_B,a_B) = U(\omega^{\text{IMU}}_B,a^{\text{IMU}}_B) + U(0,g_B) + \text{noise}. 
\label{eq:U_B}
\end{align}
We will ignore the noise terms from here on to focus on the modelling. 

In the scenario considered the reference frame $\frameA$, attached to the Earths surface, is also moving, rotating, and subject to gravity. 
For the case where the reference frame is fixed to the surface of the earth then $\omega_A = \Omega_E$ is the known Earth rotation (expressed in the reference frame $\frameA$). 
Similarly, the gravitational field $g_A$ and the proper acceleration $a^\text{p}_A$ at a fixed point on the Earths surface are also known a-priori. 
We discuss these terms in detail in \S\ref{sec:gravity_centripetal} below.
For the moment, one can write 
\begin{align} 
U(\omega_A,a_A) = U(\Omega_E,a^{\text{p}}_A) + U(0,g_A). 
\label{eq:reference_kinematics}
\end{align} 

Recalling \eqref{eq:IsochronousKinematics_3} the kinematics of $F = \idx{F}{A}{}{B}$ are written 
\begin{align}
\dot{F} & = FN - N F - U(\Omega_E,a^{\text{p}}_A)F + F U(\omega^{\text{IMU}}_B,a^{\text{IMU}}_B)\notag \\
& \makebox[1.5cm]{}- U(0,g_A)F + 
F U(0,g_B) 
\label{eq:IsochronousKinematics_v4}
\end{align}
where $\{\omega_A, a^{\text{p}}_A, g_A, g_B\}$ are known quantities, or functions of the state, and $(\omega^{\text{IMU}}_B,a^{\text{IMU}}_B)$ are measurements provided by a strap down IMU on the moving vehicle. 

\subsection{Modelling Gravity and Centripetal Force on the Earth Surface}
\label{sec:gravity_centripetal}

For a frame on the surface of the Earth, the gravitational acceleration $g_A$ points directly towards the centre of mass\footnote{
For more complex models of the Earth geoid (that model mass distribution of the Earth for example) the gravitational field may not point directly at the centre of the Earth.  However, since the associated Earth geoid is defined to balance normal and gravitational forces at the surface of the Earth, the analysis provided is holds in such a case as well.}
of the Earth as shown in Figure \ref{fig:geoid}. 
The term $a^\text{n}_A$ (normal acceleration) is associated with the contact (specific) force of the surface of the Earth acting on an object sitting at the origin of the reference frame. 
The Earth is shaped like an oblate spheroid, the \emph{geoid}, that balances gravitational and centripetal forces on the Earths surface such that the resulting contact force lies purely in the normal direction to the geoid.
Specifically, the contact force of objects resting on the earth surface is a pure normal acceleration $a_A^{\text{n}} = -g_0 \eb_3$ where $g_0 \in \R$ is the scalar gravitational constant (at the point $\idx{p}{}{}{A}$) and $\eb_3$ is the unit vector with a 1 in the third direction of an Earth fixed frame attached to the spheroid. 
The negative of the normal force is termed the \emph{apparent gravitational acceleration} $g_0 \eb_3$ and is the model for gravity used in most robotics applications.
It is only at the poles or the equator of the Earth, that the $\eb_3$ direction points towards the centre of the Earth. 
In Figure \ref{fig:geoid} the real specific forces are $g_A$ and $a^\text{n}_A$ drawn in black.  
These real specific forces are resolved into the centripetal acceleration $a^\text{c}_A$ and the apparent gravitational acceleration $g_0 \eb_3$. 
The gravitational constant varies from roughly $g_0 \sim$9.780m/s$^2$ at the Equator to about $g_0 \sim$9.832m/s$^2$ at the poles (an object will be perceived to weigh approximately 0.5\% more at the poles than at the Equator) due to centrifugal acceleration.
The proper acceleration $a_A^{\text{p}} = - g_0 \eb_3 = a_A^\text{n}$ is the specific force that would be measured by a strap down IMU placed stationary on the surface of the Earth, at the origin of the reference frame. 
Note that $a^\text{p}_A \not= -g_A$ due to the centripetal specific forces present on the surface of the Earth. 
Substituting $a^\text{p}_A = - g_0 \eb_3$ in \eqref{eq:IsochronousKinematics_v4} one has 
\begin{align}
\dot{F} & = FN - N F - U(\Omega_E,-g_0 \eb_3)F + F U(\omega^{\text{IMU}}_B,a^{\text{IMU}}_B)\notag \\
& \makebox[1.5cm]{}- U(0,g_A)F + F U(0,g_B) 
\label{eq:IsochronousKinematics_v5}
\end{align}
These equations hold anywhere on the Earth, or indeed in the solar system, as long as the gravity model $g_B = g_B(p)$ is correct (and the inertial velocity stays well below the speed of light). 

\begin{figure}
\centering
\input{figures/tikz/earth_geoid_tikz_RAM}
\caption{Force balance describing the Earth geoid. 
The geoid is defined by the balance between gravitational acceleration $g_A$ and normal acceleration $a_A^\text{n}$ perpendicular to the surface of the geoid to generate the centripetal acceleration $a_A^\text{c} = g_A + a_A^\text{n}$ required to keep a point on the surface of the geoid circling the rotational axis of the earth.  
The proper acceleration $a_A^\text{p} = a_A^\text{n}$ measured by the IMU is the normal acceleration.
The perceived gravitational acceleration is  $g_0 \eb_3 = - a_A^\text{n}$. 
}
\label{fig:geoid}
\end{figure}

As a final observation, consider the case where the vehicle considered remains close to the origin of $\frameA$, a common scenario in robotics applications. 
In this case, the variation in the gravitational field is negligible and 
\begin{align}
Rg_B(p) \approx g_A. 
\label{eq:gravity_cancellation}
\end{align}
Furthermore, $\Omega_E \approx 0$ is small and can often be ignored in many small scale robotics problems. 
In this case, expressing \eqref{eq:IsochronousKinematics_v5} in coordinates one recovers the well known kinematic model 
\begin{subequations}\label{eq:classical_kinematics}
\begin{align}
\dot{R} & = R (\omega_B^\text{IMU})^\wedge  \label{eq:classical_kinematics_R}\\
\dot{v} & = g_0 \eb_3 + R a_B^\text{IMU} 
\label{eq:classical_kinematics_v}\\
\dot{p} & = v \label{eq:classical_kinematics_p}
\end{align}
\end{subequations}
Note that only the apparent gravitational acceleration remains in the model, the true gravity fields have been cancelled, and the coordinate velocity equals the inertial velocity since the rotation of the base frame is ignored.
It is, however, important to remember that this approximation depends on the moving frame remaining close to the Earth fixed reference frame. 
As a vehicle moves away from the reference frame $\frameA$, the curvature of the Earth will begin to influence the direction of $g_B(p)$ and the model \eqref{eq:classical_kinematics_v} will no longer hold. 
For example, if a vehicle travels 100km from the reference there will be a roughly 1/60 rad angle disparity in $g_B$ and $g_A$ leading to an acceleration error (bias) of $\sim$0.15m/s$^2$ if \eqref{eq:classical_kinematics_v} is used naively. 
Similarly, as a vehicle moves far from the reference, $\Omega_E \times p$ will become significant even through $\Omega_E$ is small. 
For example, for a vehicle 100km away from the reference then $p \approx 10^{5}$m and since $\Omega_E \approx 7.29\text{\scriptsize 10}^{-5}$rad/s, the term $\Omega_E \times p$ is of order of $7$m/s (depending on the relative orientation of $p$ and $\Omega_E$). 
These errors are sufficiently important to cause significant errors in the model \eqref{eq:classical_kinematics} for systems that travel distances in the tens of kilometers. 

\section{Pose Estimation with Time-Lag and Uncertainty in Measurement Timestamps}
\label{sec:Fusion}

One of the most exciting possibilities that follow from the Galilean modelling framework is associated with the fact that Galilean frame coordinates naturally incorporate time.
This, in turn, enables temporal uncertainty to be explicitly modelled in the same framework as uncertainty in position, velocity and orientation; something that is difficult to achieve in classical pose based formulations. 
The integration of time into state estimation has already led to advances in pre-integration \cite{Delama_2024} and time-lag estimation \cite{2026_Delama} and, the authors believe, has potential to lead to advances in many robotics problems.

To illustrate the value of this approach in a robotics context, we consider the problem of fusing sensor data with uncertain timestamps into a system's state estimate. 
For example, in an aided inertial navigation system (INS), measurements from a global navigation satellite system (GNSS) receiver are fused with state estimates obtained by integrating inertial measurement unit (IMU) angular rate and acceleration data.
In practice, both the GNSS measurement timestamps and the state estimate timestamps are subject to uncertainty.
The INS fusion problem is only one of a host of robotics problems that depend on asynchronous multi-rate measurements and state updates with uncertain timestamps (consider, e.g.,~\cite{2024_Shalaby_Multi-robot}).

\subsection{Temporal Uncertainty and Galilean Frames}
\label{sec:Galilean_temporal_uncertainty} 

Consider a scenario in which the state of a robotic vehicle (defined by a Galilean frame) is estimated iteratively in real time from a sequence of measurements.
Let $k = 1, 2, \ldots$ denote the index of the sequence of state estimates. 
Denote the state of the robot at index $k$ by $F_{t_k} = (R_k,v_k,p_k, t_k)$ as a Galilean frame
\begin{equation*}
F_{t_k} = 
\begin{pmatrix}
R_k & v_k & p_k \\ 0 & 1 & t_k \\ 0 & 0 & 1
\end{pmatrix}.
\end{equation*}
with respect to a reference frame $\frameA$.
Note that the index $k$ does not explicitly correspond to a specific time, rather it is a deterministic value (i.e.~index) associated with the iterative loop of the state estimation algorithm, while the corresponding timestamp is included in the definition of the Galilean frame.

Let $F_k$ denote the information state for $F_{t_k}$ based on prior information and all measurements available to the state-estimation algorithm up to time $k$. 
We use the structure of a concentrated Gaussian discussed earlier in \S\ref{sec:Frame-uncertainty} to encode uncertainty in the information state of the system 
\begin{align}
F_k = \hat{F}_{\hat{t}_k}\exp(\epsilon_k^\wedge), \quad \epsilon_k \sim \GP(0, Q_k),
\label{eq:F_RV}
\end{align}
where $\hat{F}_{\hat{t}_k} =  (\hat{R}_k, \hat{v}_k, \hat{p}_k, \hat{t}_k) \in \Gal(3)$ is the state estimate parameter and $Q_k$ is a positive definite state covariance matrix. 
Remember that, thanks to the use of the Galilean modeling framework, the noise vector $\epsilon_k \in \mathbb{R}^{10}$ includes the uncertainty in the timestamp $t_k$ (corresponding to the $\kappa$ entry in \eqref{eq:wedge_operator}) and this uncertainty can be correlated with other states through the off diagonal elements in $Q_k$.

Consider a position measurement such as would be obtained from a GNSS receiver.
That is, consider a measurement
\begin{equation}
\label{eq:y}
y_i = p_i + \mu^p_i, \quad \mu^p_i \in \GP(0,\Sigma_i^p)
\end{equation}
taken at time $t_i$, with covariance $\Sigma_i^p \in \R^{3 \times 3}$.
The measurement has a timestamp
\begin{equation}
\label{eq:tau}
\tau_i = t_i + \mu^t_i, \quad \mu^t_i \in \GP(0,\Sigma_i^t),
\end{equation}
where $\Sigma_i^t \in \R$ is the uncertainty in the timestamp, often referred to as jitter.\footnote{
In practice, the noise associated with timestamps is not usually Gaussian, however, it is common to assume Gaussian jitter (around a known delay) to simplify algorithm development. 
}
The measurement index $i$ is deterministic.
The index $i$ and the index $k$ have no direct connection and will have different timestamps (in general) and often be available at different rates.
Typically, for a GNSS system the measurement timestamp $t_i$ is delayed with respect to the system timestamp $t_i < t_k$ due to the processing time of the GNSS system. 
The model \eqref{eq:tau} assumes that the mean delay is known, or at least an unbiased estimate is available~\cite{2026_Delama}, and the uncertainty (jitter) is incorporated into the timestamp noise $\Sigma_i^t$.
Define $y_{\tau_i} = (y_i, \tau_i) \in \R^4$ to be the event associated with the measurement and its time stamp. 

Let $\ob{y}_0 := \ob{0}_0 = (0, 0,0,0,1)^\top$ denote the zero position at reference time zero in homogeneous coordinates.
Let $\Sigma_{i} = \diag(\Sigma_i^p, \Sigma_i^t) \in \R^{4 \times 4}$ be the combined measurement covariance. 
Recall that the homogeneous coordinates of event noise are $\ob{\mu}_i = (\mu^p_i, \mu_i^t, 0)$~\eqref{eq:homog_free_event}, then the homogeneous coordinates for the measurement event
\begin{align}
\ob{y}_{\tau_i} = 
\begin{pmatrix} 
y_i \\ \tau_i \\ 1 
\end{pmatrix}
& = F_{t_i} \ob{y}_0 + \ob{\mu}_i, \qquad \mu_i \sim \GP(0,\Sigma_i),
\label{eq:ybar_model}
\end{align}
where $F_{t_i} = (R_i,v_i,p_i,t_i)$ is the true state of the vehicle at time $t_i$ of the measurement. 

\subsection{Pre-integration}
\label{sec:PreIntegration}

\nocite{barrau2020mathematical}

\begin{figure}
\centering
\includegraphics[scale=0.45]{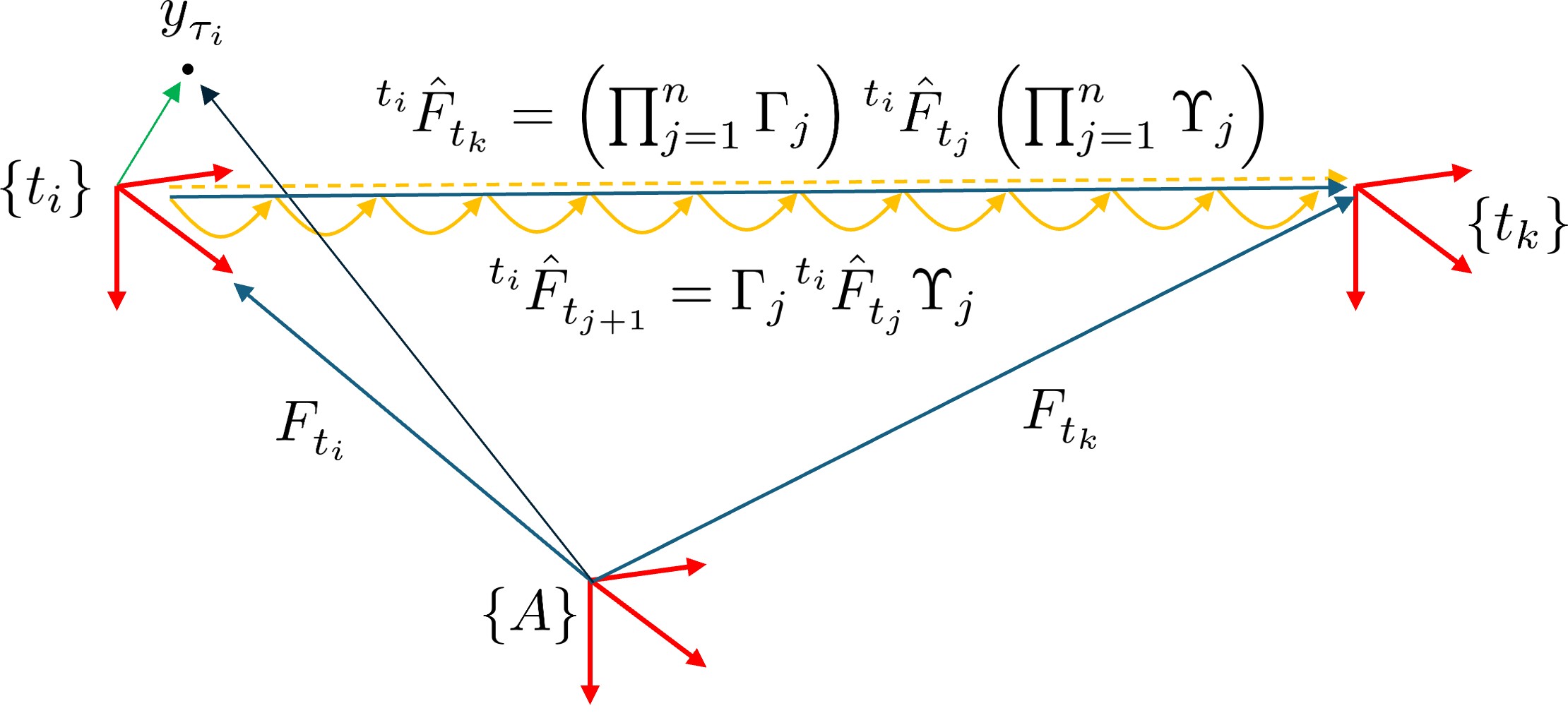}
\caption{
The measurement $y_{\tau_i}$ is associated with frame $\{t_i\}$ with Galilean coordinates $F_{t_i}$ with respect to a reference $\frameA$. 
The goal is to filter for the state $F_k$ associated with the frame $\{t_k\}$. 
The relationship between $\{t_i\}$ and $\{t_k\}$ is given by the Galilean frame $\idx{F}{t_i}{}{t_k} =\idx[-1]{F}{}{}{t_i} \idx{F}{}{}{t_k}$ that can be computed by pre-integrating IMU data to compute $F_t$ over the time interval $t \in [t_i , t_k]$.
Each small hop corresponds to an IMU measurement update $\idx{F}{t_i}{}{t_j} \mapsto \Gamma_j \idx{F}{t_i}{}{t_j} \Upsilon_j$ \eqref{eq:preintegration_Euler}.
} 
\label{fig:delayedMeasurement}
\end{figure}

The power of the Galilean formulation lies in the ability to reconcile states and measurements taken at different times.  
Consider the measurement event $\ob{y}_{\tau_i}$ and note that it is expressed with respect to a reference frame $\frameA$ (the black arrow in Fig.~\ref{fig:delayedMeasurement}). 
The model \eqref{eq:ybar_model} is defined with respect to the Galilean frame $F_{t_i}$ where the green arrow in Figure~\ref{fig:delayedMeasurement} represents the spatio-temporal noise $\mu_i$ in \eqref{eq:ybar_model}. 
We need to derive a model for $\ob{y}_{\tau_i}$ in terms of $F_{t_k}$, since the associated information state $F_k$ is the variable updated in the fusion step. 
One can write 
\begin{align*}
\ob{y}_{\tau_i} = & F_{t_k} F_{t_k}^{-1} F_{t_i} \ob{y}_0 + \ob{\mu}_i \\ 
= & F_{t_k} \idx[-1]{F}{t_i}{}{t_k} \ob{y}_0 + \ob{\mu}_i,
\end{align*}
where $\idx{F}{t_i}{}{t_k} = F_{t_i}^{-1} F_{t_k}$ is the Galilean frame coordinates of $\{t_k\}$ expressed with respect to $\{t_i\}$ as shown in Figure \ref{fig:delayedMeasurement}. 
The transformation $\idx{F}{t_i}{}{t_k}$ is analogous to a pre-integration transformation \cite{lupton2011visual,forster2016manifold}, expressed as a Galilean transformation \cite{Delama_2024}, since it is associated with integrating the motion of the vehicle to propagate the measurement event at time $t_i$ forward to time $t_k$ before incorporating it in the state-estimation process. 
In order to use this formula in the fusion algorithm one needs to generate an estimate $\idx{\hat{F}}{t_i}{}{t_k}$ of the Galilean frame $\idx{F}{t_i}{}{t_k}$. 

In common with most robotics scenarios, we assume that there are IMU measurements available that can be integrated to provide an estimate of $\idx{F}{t_i}{}{t_k}$. 
We will also assume that the system is subject to gravity but we will ignore any rotation effects of an Earth fixed reference in common with the majority of robotics treatments. 

Fix $\{t_i\}$ as a reference frame and consider the continuous kinematics of the frame $\idx{F}{t_i}{}{t}$ for $t \in [t_i, t_k]$. 
Recalling \eqref{eq:isochronous_inertial_Galilean_kinematics} one has  
\begin{align} 
\idx{\dot{F}}{t_i}{}{t} 
& = \idx{F}{t_i}{}{t} \,\left( \idx{U}{}{}{t} + N + G_B \right), 
\quad\quad\quad  
\idx{F}{t_i}{}{t_i} = \Id_5
\label{eq:Galilean_pre-integration} 
\end{align}
where $\idx{U}{}{}{t} = U(\omega_t, a_t)$ are the inertial IMU measurements \eqref{eq:kinematics_Galilean}, $N$ captures the linear kinematics \eqref{eq:N}, and $G_B = U(0, R_{t_i}^\top \idx{g}{}{}{B} )$ captures the gravitational input in the body-frame, analogous to \eqref{eq:U_B}.

Let $\idx{g}{}{}{A}$ denote the known (constant) local gravitational model in the reference frame $\frameA$. 
Define $\idx{g}{}{}{i} := \idx{R}{t_i}{}{A} g_A$ to be the gravitational direction in the $\{t_i\}$ frame. 
Note that for the purposes of the pre-integration, $\idx{g}{}{}{i}$ is also a constant vector. 
The body gravity coordinates 
\[
\idx{g}{}{}{B} = \idx{R}{t}{}{A} \idx{g}{}{}{A}
= \idx{R}{t}{}{t_i} \idx{R}{t_i}{}{A} \idx{g}{}{}{A}
= \idx{R}{t}{}{t_i} \idx{g}{}{}{i}
\]
are time varying and not known a-priori since $\idx{F}{t_i}{}{t}$, and hence $\idx{R}{t}{}{t_i} = \idx[\top]{R}{t_i}{}{t}$, is not known.
However, one can shift the gravity term $G_B$ in \eqref{eq:Galilean_pre-integration} from a left invariant to a right invariant vector field structure. 
Considering just the gravity term from \eqref{eq:Galilean_pre-integration}, one has 
\begin{align*}
\idx{F}{t_i}{}{t} G_B & = 
\left( \idx{F}{t_i}{}{t} G_B \idx[-1]{F}{t_i}{}{t} \right)  \idx{F}{t_i}{}{t} = 
( G^a_i - (t - t_i) G^v_i ) \idx{F}{t_i}{}{t} 
\end{align*}
where 
\[
G^a_i = \begin{pmatrix} 0 & \idx{g}{}{}{i} & 0   \\ 0 & 0 & 0 \\ 0 & 0 & 0 \end{pmatrix},   \text{ and } 
G^v_i = \begin{pmatrix} 0 & 0 & \idx{g}{}{}{i}   \\ 0 & 0 & 0 \\ 0 & 0 & 0 \end{pmatrix} 
\]
capture the corresponding effect of gravity as a left input in frame $\{t_i\}$. 
The advantage of this manipulation is that while $G_B$ depends on the state $\idx{F}{t_i}{}{t}$, the terms $G^a_i$ and $G^v_i$ are known constant matrices. 
It follows that one may write kinematics for the estimated state $\idx{\hat{F}}{t_i}{}{t}$ as 
\begin{align} 
\idx{\dot{\hat{F}}}{t_i}{}{t} = & ( G^a_i - (t - t_i) G^v_i ) \idx{\hat{F}}{t_i}{}{t} +  \idx{\hat{F}}{t_i}{}{t} \,\left( \idx{U}{}{}{t} + N\right), 
\label{eq:time_offset_kinematics} \\ 
& \qquad\qquad\qquad\qquad\qquad\qquad\qquad\qquad 
\idx{\hat{F}}{t_i}{}{t_i} = I_5. \notag
\end{align} 
Compare this equation to \eqref{eq:IsochronousKinematics_v5} where the reference was isochronous.
Here we fix the reference $\{t_i\}$, a Galilean reference, rather than the kinematics \eqref{eq:reference_kinematics} of an isochronous reference, and there is no gravity cancellation \eqref{eq:gravity_cancellation}. 
When the body gravity is transformed into a left invariant input it gains position kinematics $\dot{p} = -(t-t_i)\idx{g}{}{}{i}$, in addition to the spatial gravity kinematics $\dot{v} = \idx{g}{}{}{i}$, that compensate for the fixed Galilean reference frame. 

Consider the case where $n$ IMU measurements $(\omega_{t_j}, a_{t_j})$ are available at a sequence of times $t_j \in [t_i, t_k)$ for $j = 1, \ldots, n$.
Applying a zero-order hold assumption we assume the input $\idx{U}{}{}{t} = U(\omega_{t_j}, a_{t_j})$ is constant for the period $t \in [t_j,t_{j+1})$. 
Since $G^a_i G^v_i = G^v_iG^a_i$ commute it is possible to solve \eqref{eq:time_offset_kinematics} (with initial condition $\idx{\hat{F}}{t_i}{}{t_j}$) for $t \in [t_j,t_{j+1})$ analytically  
\begin{align}
\idx{\hat{F}}{t_i}{}{t_{j+1}} & = 
\exp\left( (t_{j+1} - t_j)G^a_i - \frac{(t_{j+1}-t_j)^2}{2} G^v_i \right) \cdot 
 \label{eq:preintegration_Euler}
\\ 
& \qquad\qquad\qquad\quad
\idx{\hat{F}}{t_i}{}{t_j} \exp \left((t_{j+1} - t_j) (\idx{U}{}{}{t_j} + N) \right). \notag
\end{align}
It follows that the integral of \eqref{eq:Galilean_pre-integration} can be approximated by a sequence of small hops (cf.~Fig.~\ref{fig:delayedMeasurement}) $\idx{F}{t_i}{}{t_{j+1}} = \Gamma_j \idx{F}{t_i}{}{t_j} \Upsilon_j$ where 
\begin{align}
\Gamma_j & = \exp \left( (t_{j+1} - t_j )G^a_i - \frac{(t_{j+1}-t_j)^2}{2} G^v_i \right),  \label{eq:Upsilon}\\ 
\Upsilon_j & = \exp \left((t_{j+1} - t_j) (\idx{U}{}{}{t_j} + N) \right). \label{eq:Gamma}
\end{align}
Assume that the limiting times are $t_{[j = 1]} = t_i$ and $t_{[j = n+1]} = t_k$ (cf.~Fig.~\ref{fig:delayedMeasurement}) then 
\begin{align}
\idx{\hat{F}}{t_i}{}{t_{k}} & \approx (\prod_{j=1}^n \Gamma_j) \idx{\hat{F}}{t_i}{}{t_i}  (\prod_{j=1}^n \Upsilon_j).
\label{eq:preintegration}
\end{align}

In conclusion, recalling that $\idx{\hat{F}}{t_k}{}{t_i} = \idx[-1]{\hat{F}}{t_i}{}{t_k}$, one can write 
\begin{align}
\ob{y}_{\tau_i} 
= & F_{t_k} \idx{\hat{F}}{t_k}{}{t_i} \ob{y}_0 + \ob{\mu}'_i, \qquad \mu'_i \sim \GP(0,\Sigma'_i)
\label{eq:oby_taui}
\end{align}
as a function of $F_{t_k}$. 
Here $\Sigma'_i$ is a modified noise process, augmented to take account of uncertainty in $\idx{\hat{F}}{t_i}{}{t_k}$ due to noise in the IMU measurements. 
For this tutorial paper, we have chosen to leave out the details of estimating this uncertainty, since the computations are somewhat technical, tedious, and provide very little additional intuition.
The interested reader is referred to \cite{2026_Delama}.

\subsection{Error Coordinates}
\label{sec:Error_Coordinates}

Rather than working directly with the variables $F_{t_k}$, $\hat{F}_{\hat{t}_k}$, and $\ob{y}_{\tau_i}$, it is advantageous to transform the problem into error coordinates \cite{1982_Maybeck_stochastic_book}, \cite{1999_Roumeliotis_errorstate}, \cite{2018_Carpenter_navigation}, \cite{2023_vanGoor_EqF} around the state estimate $\hat{F}_{\hat{t}_k}$. 
A key advantage is that the corresponding Baysian fusion problem can be formulated as an optimisation on the Lie algebra $\gal(3)$. 
For estimation problems on groups, the error coordinates allow the lift and retraction between group and tangent space in a well defined area around the origin and inherently avoid domains where exponential map is ill conditioned or not defined and potential singularities, improving consistency of the resulting filtering algorithms. 
This approach minimizes linearisation error that would be incurred if a single logarithmic (or any other) chart is used to define coordinates on the group. 
Furthermore, for groups, the definition of the error state around the origin directly supports the structure of concentrated Gaussians \cite{Wang_2006_Error,2024_Ge_LCSS} to represent the error state uncertainty (see Section \S\ref{sec:Frame-uncertainty}). 

Recall \eqref{eq:F_RV} and define a new error random variable \cite{2023_vanGoor_EqF}
\begin{equation}
E_k = \hat{F}_{\hat{t}_k}^{-1} F_k  = \exp(\epsilon_k^\wedge), \quad \epsilon_k \sim \GP(0,Q_k).
\label{eq:prior}
\end{equation}
The error variable $E_k$ represents the relative Galilean reference frame of $F_{t_k}$ with respect to $\hat{F}_{\hat{t}_k}$.
Similarly, the measurement event $\ob{y}_{\tau_i}$ is transformed into an event $\idx{\ob{d}}{}{}{i}$ relative to the frame $\hat{F}_{\hat{t}_k}$; 
\begin{align}
\idx{\ob{d}}{}{}{i} & = \hat{F}_{\hat{t}_k}^{-1} \ob{y}_{\tau_i}, \qquad \mu'_i \sim 
\GP(0, \Sigma'_{i})\notag  \\
& = 
\hat{F}_{\hat{t}_k}^{-1} F_{t_k} \idx[-1]{\hat{F}}{t_i}{}{t_k} \ob{y}_0 + \hat{F}_{\hat{t}_k}^{-1}  \ob{\mu}'_i,  \notag  \\
& = 
E_k \idx[-1]{\hat{F}}{t_i}{}{t_k} \ob{y}_0 + \idx{\ob{\mu}}{k}{}{i}, 
\quad \idx{\mu}{k}{}{i}\sim \GP(0, \idx{\Sigma}{k}{}{i}) 
\label{eq:d_ki} 
\end{align}
where the second line follows from substituting for \eqref{eq:oby_taui} and the covariance 
\[
\idx{\Sigma}{k}{}{i} = 
\left[ \hat{F}_{\hat{t}_k}^{-1} 
\ob{\Sigma}'_i
\hat{F}_{\hat{t}_k}^{-\top}
\right]_{[1:4,1:4]}
\]
is straightforward to compute since the action $\hat{F}_{\hat{t}_k}^{-1} \ob{\mu}'_i$ is linear. 
Here $\ob{\Sigma}'_i$ is the 5$\times$5 matrix with $\Sigma'_i$ in the upper left 4$\times$4 block and zeros in the bottom row and last column. 
The subscript ${[1:4,1:4]}$ indicates you take the upper left 4$\times$4 block of the matrix in the braces.
The event $\idx{\ob{d}}{}{}{i}$ is termed the \emph{equivariant innovation} \cite{2021_mahony_HomogeneousSpaceGeometry} since it plays the same role as the innovation $\tilde{y} = y - \hat{y}$ in classical fusion algorithms but is constructed through the application of a group action.

\subsection{Bayesian Fusion}
\label{sec:Baysian_Fusion}

The fusion problem in error coordinates is to approximate the information state by a concentrated Gaussian.
That is, to compute the maximum a posteriori (MAP) estimate of $E_k$ given $\idx{d}{}{}{i}$ and the prior distribution \eqref{eq:prior}.
For the purposes of this discussion we will assume that the prior for $E_k$ and the generative measurement noise for $\idx{d}{}{}{i}$ are independent stochastic process.
By Bayes' rule, the posterior distribution is
\begin{equation}
\label{eq:fusion_equation}
p(E_{k} \mid \idx{d}{}{}{i})
= \frac{1}{\beta}\,p(\idx{d}{}{}{i} \mid E_k)\,p(E_{k}),
\end{equation}
where $\beta = p(\idx{d}{}{}{i})$ is a normalising constant that is independent of the error $E_k$.

There are a number of approaches to solving \eqref{eq:fusion_equation} that are well documented in the literature \cite{barfoot2024state}.
The most common method is to treat $\epsilon_k$ as local log coordinates for the posterior estimate of $E_k$ around the origin $I_5$,
\begin{equation}
\label{eq:E_k}
\epsilon_k^\wedge = \log (E_k), \quad \epsilon_k \in \R^{10} 
\end{equation}
where here the $\log$ is the $\Gal(3)$ log function. 

Define 
\begin{align}
\ob{I}_4 := \begin{pmatrix} 1 & 0 & 0 & 0 \\ 0 & 1 & 0 & 0 \\ 0 & 0 & 1 & 0 \\ 0 & 0 & 0 & 1 \\ 0 & 0 & 0 & 0   \end{pmatrix} 
\label{eq:obI_4}
\end{align}
in order that $\ob{I}^\top \idx{\ob{d}}{}{}{i} = \idx{d}{}{}{i}  \in \R^4$ extracts the event coordinates from event homogeneous coordinates. 
Substituting $E_k = \exp(\epsilon_k^\wedge) \approx I_5 + \epsilon_k^\wedge$ yields the residual
\begin{align}
\label{eq:residual}
(\idx{d}{}{}{i} - \ob{I}_4^\top E_k \idx{\hat{F}}{t_k}{}{t_i} \ob{y}_0)
& \approx
(\idx{d}{}{}{i} - \ob{I}_4^\top \idx{\hat{F}}{t_k}{}{t_i} \ob{y}_0) 
- \ob{I}_4^\top\epsilon_k^\wedge \idx{\hat{F}}{t_k}{}{t_i} \ob{y}_0 \\
& = 
\idx{\tilde{d}}{}{}{i} - \ob{I}_4^\top \epsilon_k^\wedge \idx{\hat{F}}{t_k}{}{t_i} \ob{y}_0,
\end{align}
where
\vspace{-0.5\baselineskip}
\begin{align}
\label{eq:d_tilde}
\idx{\tilde{d}}{}{}{i} 
= \idx{d}{}{}{i} - \ob{I}_4^\top  \idx{\hat{F}}{t_k}{}{t_i} \ob{y}_0
\end{align}
Note that $\epsilon_k \mapsto \ob{I}_4^\top \epsilon_k^{\wedge}
 \idx{\hat{F}}{t_k}{}{t_i} \ob{y}_0$ is a linear function of $\epsilon_k$.
Define the linear matrix $\idx{C}{}{}{i} \in \R^{4 \times 10}$ such that 
\begin{align}
\idx{C}{}{}{i} \epsilon_k :=  \ob{I}_4^\top \epsilon_k^{\wedge} \idx{\hat{F}}{t_k}{}{t_i} \ob{y}_0.
\label{eq:C}
\end{align}
From this point, then one can follow the classical Kalman filter process, optimise the log-likelihood of \eqref{eq:fusion_equation} and obtain the mean $\hat{\epsilon}_k$ of the information error state distribution in local coordinates
\begin{equation*}
\hat{\epsilon}_k = (Q_k^{-1} + \idx{C}{}{}{i}^\top \idx[-1]{\Sigma}{k}{}{i} \idx{C}{}{}{i} )^{-1} \idx{C}{}{}{i}^\top \idx[-1]{\Sigma}{k}{}{i}\idx{\tilde{d}}{}{}{i},	
\end{equation*}
where $(Q_k^{-1} + \idx{C}{}{}{i}^\top \idx[-1]{\Sigma}{k}{}{i} \idx{C}{}{}{i} )$ represents the innovation matrix corresponding to the residual.

Recalling \eqref{eq:E_k}, the posterior mean of the original fusion problem $p(F_k \mid \ob{y}_{\tau_i}, \hat{F}_{\hat{t}_k})$ is given by
\[
F^\ast_k = \hat{F}_{\hat{t}_k} \exp(\hat{\epsilon}_k^{\wedge}).
\]

For the covariance estimate, one may use the completion of the squares formulae directly to estimate the covariance.
However, a slightly better approximation is provided by noting that the covariance is the Hessian of the log-likelihood function at $\hat{\epsilon}_k$.
Since the log-likelihood function $p(\idx{d}{}{}{i} \mid E_{k})$ is actually a function of $E_k$ and not $\epsilon_k$ (see also approximation in (\ref{eq:residual})), one can account for the nonlinearity of the local coordinates by including the `right' \cite{chirikjian2009stochastic} Jacobian of the exponential map $J_{\hat{\epsilon}_k}^R$, evaluated at the point $\hat{\epsilon}_k$, in the covariance estimate
\begin{align*}
Q^\ast_k = 
(J^R_{\hat{\epsilon}_k})^\top (Q^{-1}_k + \idx{C}{}{}{i}^\top \idx[-1]{\Sigma}{k}{}{i} \idx{C}{}{}{i})^{-1} 
J^R_{\hat{\epsilon}_k}.	
\end{align*}
Analytic expressions for the Jacobian $J^R_{\hat{\epsilon}_k}$ of the Galilean group have been published recently in \cite[Eq.~(10)]{Delama_2024} see also \cite{barfoot2025integral,2024_Kelly_arxiv}.
Including the Jacobian is equivalent to including the reset step in an extended Kalman filter 
\cite{markley2003attitude,mueller2017covariance,gill2020full,ge2025geometryextendedkalmanfilters}.

\begin{figure}
\centering
\includegraphics[width=0.8\columnwidth]{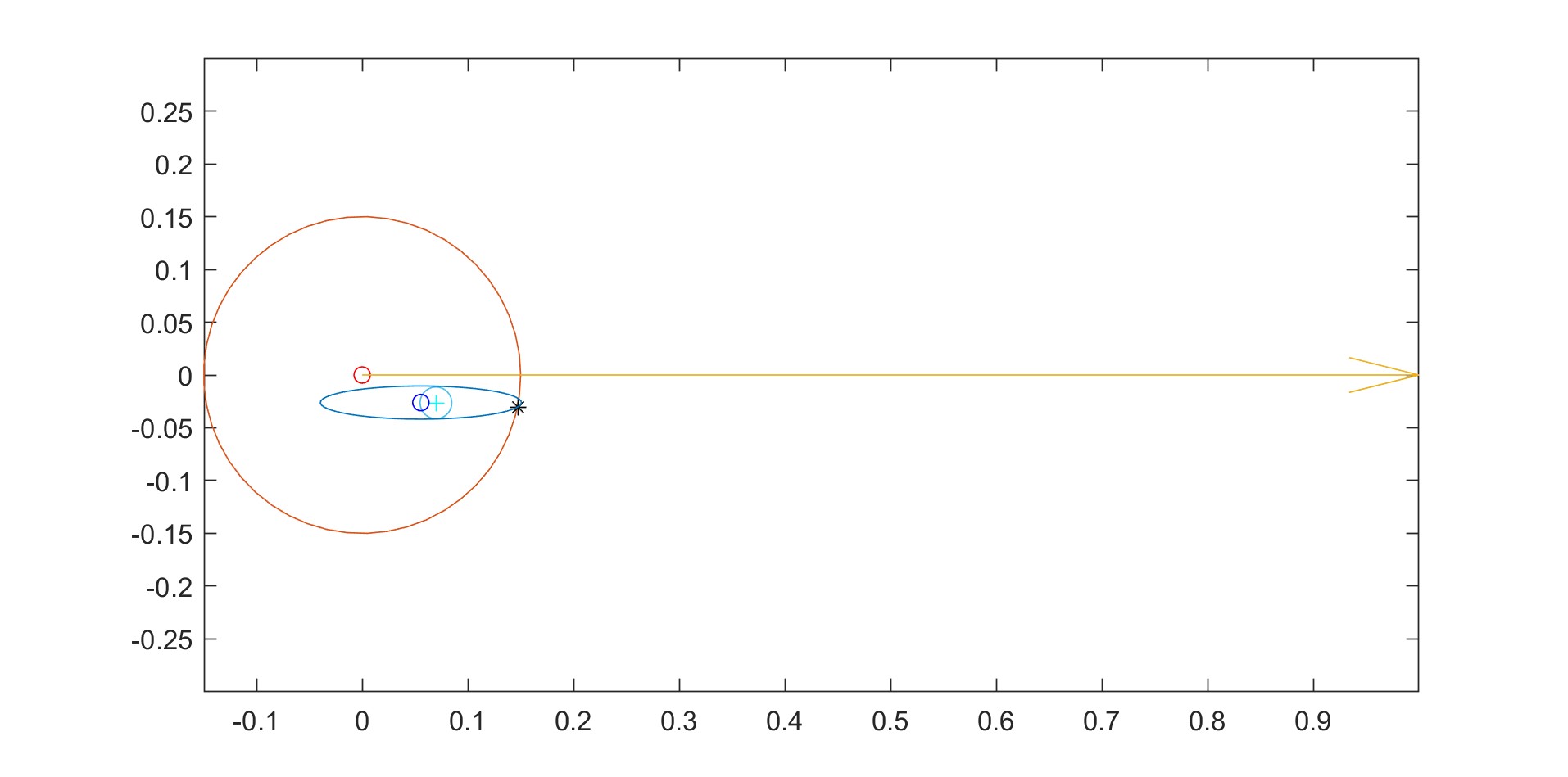}
\includegraphics[width=0.8\columnwidth]{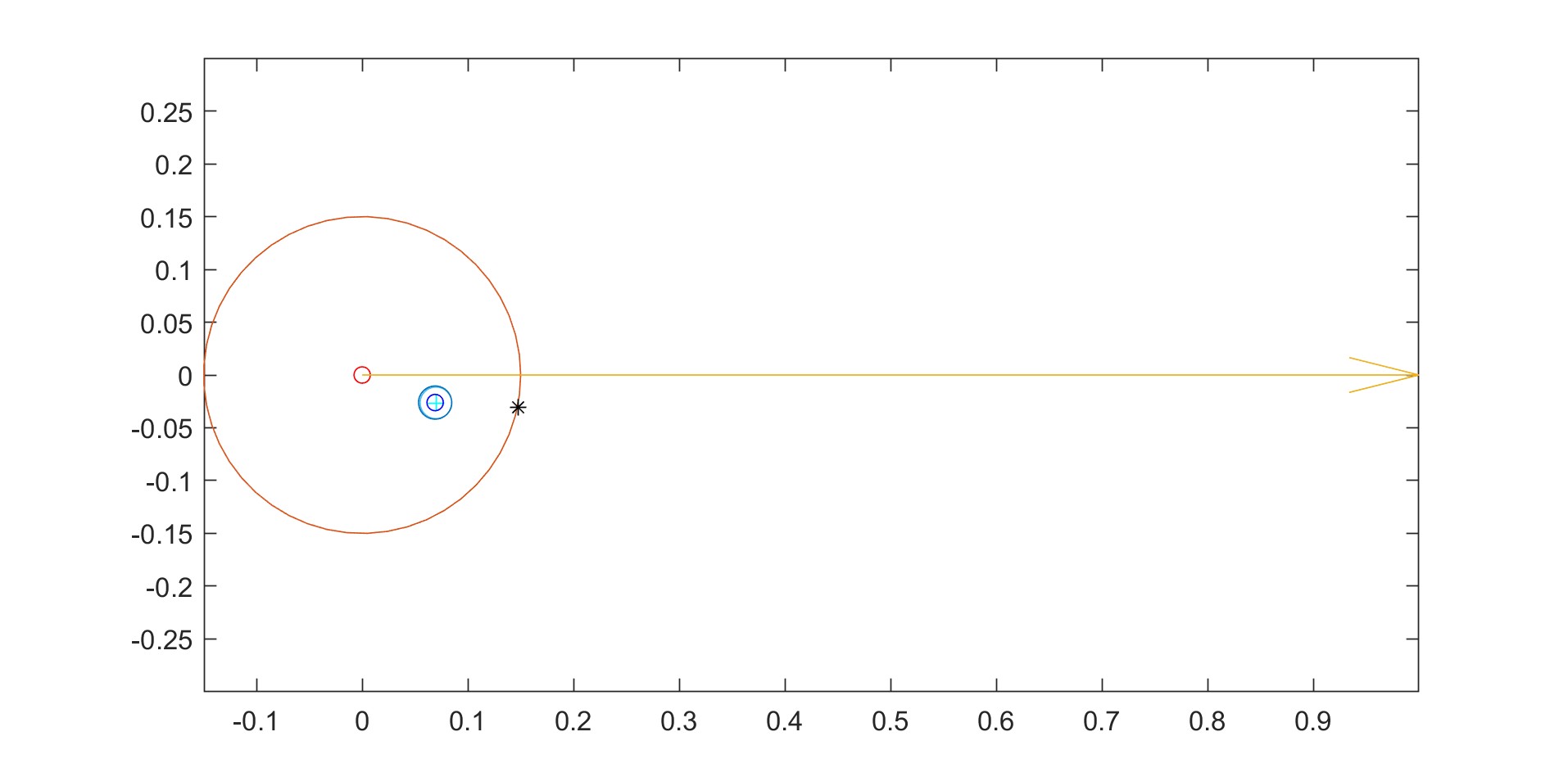}
\caption{Fusion using the Galilean construction (top figure) compared to using classical fusion (bottom figure) for the case where there is high uncertainty in the measurement timestamp. 
The prior state estimate is the small red circle (at $(0,0)$) with red homogeneous uncertainty ellipse (large red circle centred at $(0,0)$). 
The system is travelling at velocity 1m/s as shown by the arrow. 
The measurement is shown as a cyan cross with cyan homogeneous uncertainty ellipse. 
The true state is shown by the black asterisk. 
The fused estimate is the blue circle with blue uncertainty ellipse. 
Note that for the Galilean fusion the uncertainty in the temporal coordinate stretches the uncertainty of the state estimate in the direction of the velocity. 
The temporal uncertainty is not modelled in classical fusion and the posteriori state estimate is homogeneous and overconfident. }
\label{fig:fusion_sim}
\end{figure}

In conclusion, the resulting, fused information state is the concentrated Gaussian
\[
p(F_k \mid \ob{y}_{\tau_i}, \hat{F}_{\hat{t}_k}) = \GP(F^\ast_k, Q^\ast_k).
\]
A simple simulation of fusion is shown in Figure~\ref{fig:fusion_sim}. 
The optimisation will automatically trade off error in the time-stamp against error in the classical navigation states.
For example, if the GPS measurement noise lies along the velocity direction of the estimate, then the fusion will partly adjust the timestamp to allow the natural velocity dynamics to improve the error, rather than forcing all the correction into position and velocity.

This example, provides insight into the power of the Galilean modelling framework in dealing with temporal uncertainty, as well as reconciling measurements from different time-stamps using pre-integration. 
The mathematical machinery used, apart from the Galilean structure, is a direct analogue of recent work in equivariant filtering \cite{ge2025geometryextendedkalmanfilters, 2021_mahony_HomogeneousSpaceGeometry,2023_vanGoor_EqF}.  
The generalisation to the new structure is conceptually straightforward and demonstrates the power of the Galilean framework in overcoming roadblocks in a wide range of critical robotics problems \cite{kelly2014general,kelly2021question,2024_Shalaby_Multi-robot,Delama_2024,2026_Delama}. 

\section{Conclusion}
\label{sec:Conclusion}

We have presented a robotics-centric treatment of Galilean symmetry, highlighting its relevance to modern robotics problems. 
While the use of rigid-body transformations is ubiquitous in robotics, Galilean symmetry provides a broader geometric foundation that explicitly incorporates both time and motion. 
Despite its well-established role in classical physics, its application in robotics has remained largely unexplored. Our aim has been to begin to bridge this gap, making the concepts of Galilean symmetry more accessible to roboticists.
This work just scratches the surface of what Galilean symmetry can offer and the authors hope the paper inspires further investigation of this rich and elegant structure by the robotics community.

\section*{Acknowledgments}
This research was supported by the Australian Research Council
through Discovery Grant DP210102607 ``Exploiting the Symmetry of
Spatial Awareness for 21st Century Automation,'' by the Natural Sciences and Engineering Research Council of Canada through Discovery Grant RGPIN-2023-05036 ``Learning on Manifolds for Robotic Manipulation,'' and the Austrian Ministry of Climate Action and Energy (BMK) under the grant agreement 925719 (NightWatch).


 
\newcommand{\etalchar}[1]{$^{#1}$}


\end{document}

%% file: preamble_arXiv.tex
\usepackage{hyperref}
\usepackage{microtype}

\usepackage{amssymb}
\usepackage{amsmath}
\usepackage{amsthm}
\usepackage{mathdots}
\usepackage{mathtools}

\usepackage{tensor}
\usepackage{xifthen}

\usepackage{urwchancal}
\DeclareFontFamily{OT1}{pzc}{}
\DeclareFontShape{OT1}{pzc}{m}{it}{<-> s * [1.10] pzcmi7t}{}
\DeclareMathAlphabet{\mathpzc}{OT1}{pzc}{m}{it}

\usepackage{graphicx}
\usepackage{ifpdf}
\ifpdf
\usepackage{epstopdf}
\PrependGraphicsExtensions{.svg}
\fi

\usepackage{xypic}

\newtheorem{theorem}{Theorem}[section]

\newtheorem{remark}[theorem]{Remark}

\newcommand{\QED}{\begin{flushright}$\Box$\end{flushright}}


%% file: notation_defs.tex
\providecommand{\R}{\mathbb{R}}

\providecommand{\SO}{\mathbf{SO}}

\providecommand{\SE}{\mathbf{SE}}
\providecommand{\Gal}{\mathbf{Gal}} 

\providecommand{\grpG}{\mathbf{G}}

\providecommand{\grpR}{\mathbf{R}}

\providecommand{\gothse}{\mathfrak{se}}

\providecommand{\gothgal}{\mathfrak{gal}}

\providecommand{\se}{\mathfrak{se}}
\providecommand{\gal}{\mathfrak{gal}}

\providecommand{\Sph}{\mathrm{S}}

\providecommand{\cset}[2]{\left\{ {#1} \mid {#2} \right\}}

\providecommand{\tT}{\mathrm{T}} 

\providecommand{\Id}{I} 

\providecommand{\eb}{\mathbf{e}} 

\providecommand{\GP}{\mathbf{N}} 

\DeclareMathOperator{\diag}{diag}

\providecommand{\td}{\mathrm{d}}

\providecommand{\ddt}{\frac{\td}{\td t}}

\def\frameA{\mbox{$\{A\}$}}

\def\frameB{\mbox{$\{B\}$}}

\def\frameZero{\mbox{$\{0\}$}}

\makeatletter
\newcommand{\ubar}[1]{%
  \mathchoice
    {\ul@do\displaystyle{#1}{0.45ex}{0.1ex}}
    {\ul@do\textstyle{#1}{0.45ex}{0.1ex}}
    {\ul@do\scriptstyle{#1}{0.28ex}{0.08ex}}
    {\ul@do\scriptscriptstyle{#1}{0.20ex}{0.06ex}}
}
\newcommand{\ul@do}[4]{
  \setbox0=\hbox{$\m@th#1#2$}%
  \mathord{%
    \ooalign{%
      \copy0\cr
      \hidewidth\lower#3\hbox{\vrule height #4 width 0.7\wd0}\hidewidth\cr
    }%
  }%
}
\makeatother

\providecommand{\ob}[1]{\overline{#1}} 
\usepackage{accents}
\usepackage{mathtools}
\makeatletter
\providecommand{\scirc}{%
    \hbox{\fontfamily{\rmdefault}\fontsize{0.4\dimexpr(\f@size pt)}{0}\selectfont{\raisebox{-0.52ex}[0ex][-0.52ex]{$\circ$}}}}

\makeatother

\makeatletter
\providecommand{\ucirc}{%
    \hbox{\fontfamily{\rmdefault}\fontsize{0.4\dimexpr(\f@size pt)}{0}\selectfont{\raisebox{0.0ex}[0ex][-0.52ex]{$\circ$}}}}

\makeatother

\mathchardef\mhyphen="2D

\providecommand{\idx}[5][]{
\ifthenelse{\isempty{#1}}
{\tensor*[_{#4}^{#3}]{#2}{_{#5}}}
{\tensor*[_{#4}^{#3}]{#2}{^{#1}_{#5}}}
}

\providecommand{\etal}{\textit{et al.~}}



%% file: figures/tikz/velocities_tikz_RAM.tex
\begin{tikzpicture}[scale=4, >=stealth, line cap=round]
  \coordinate (A) at (0,0);
  \draw[->, very thick, red] (A) -- ++(0.21,0.04);
  \draw[->, very thick, red] (A) -- ++(0,-0.220);
  \draw[->, very thick, red] (A) -- ++(0.175,-0.11);
  \node[below left, xshift=-2pt, yshift=-3pt, black] at (A) {$\{A\}$};
  
  \draw[->, thick, yellow!60!red] (A) -- ++(-0.35,0.1) 
  	node[below left, xshift=10pt, yshift=-5pt, black]{$w$};

  \draw[->, thick, green!70!black] (A) -- ++(-0.068,0.29)
    node[left, xshift=-2pt, black] {$\Omega$};

  \begin{scope}[rotate around={15:(A)}]
    \draw[->, thick, blue!65!green, transform shape=false] ($(A) + (0.005,0.19)$)
      arc[start angle=-90, end angle=75, x radius=0.11, y radius=0.02];
  \end{scope}  

  \begin{scope}[rotate around={15:(A)},xscale=-1]
    \draw[thick, blue!65!green] ($(A) + (-0.005,0.19)$)
      arc[start angle=-90, end angle=75, x radius=0.11, y radius=0.02];
  \end{scope}
  
  \coordinate (P) at (0.55,0.35);
  
  \draw[->, thick, yellow!60!red] (P) -- ++(0.35,-0.1) 
  	node[above right, xshift=-10pt, yshift=2pt, black]{$v_p$};

  \draw[->, thick, green!70!black] (P) -- ++(-0.08,-0.28) 
  	node[below, xshift=-2pt, yshift=1pt, black]{$-\Omega \times p$};
  
  \draw[->, thick, cyan!80] (P) -- ++(0.18,-0.33) 
  	node[right, xshift=3pt, yshift=-0.98pt, black]{$\dot{p} = -\Omega \times p + (v_p - w)$};

  \begin{scope}[rotate around={15:(P)}]
    \coordinate (Q) at ($(P) + (-0.25, -0.035)$);
    \draw[dashed, thick, blue!65!green, transform shape=false] (Q)
      arc[start angle=-80, end angle=80, x radius=0.3, y radius=0.04];
  \end{scope}

  \draw[black] (A) -- (P);
  
  \fill (P) circle[radius=0.5pt] node[above, yshift=1pt]{$p$};
\end{tikzpicture}

%% file: figures/tikz/kinematics_tikz_RAM.tex
\begin{tikzpicture}[scale=4, >=stealth, line cap=round]
  \coordinate (O) at (0,0);
  \draw[->, very thick, red] (O) -- ++( 0.225,0);
  \draw[->, very thick, red] (O) -- ++( 0,0.225);
  \draw[->, very thick, red] (O) -- ++(-0.12,-0.12);
  \node at ($(O) + (0.080,-0.110)$) {$\{0\}$};

  \coordinate (E) at (-0.50,0.650);

  \draw[dashed, thick, black!65] (E) -- ++(0,-0.55);
  \draw[dashed, thick, black!65] (E) -- ++(0, 0.55)
    node[above left, yshift=-3pt, black] {$\Omega_E$};

  \node at (E) {\includegraphics[width=3.2cm]{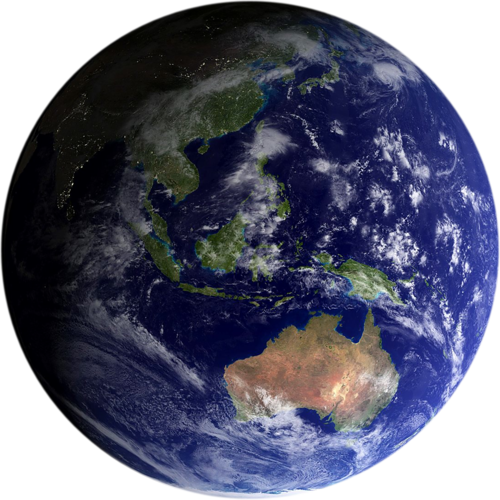}};

  \draw[->, thick, blue!65!green, transform shape=false] ($(E) + (0,0.45)$) 
    arc[start angle=-90, end angle=75, x radius=0.15, y radius=0.03];

  \begin{scope}[xscale=-1]
    \draw[thick, blue!65!green] ($(E) + (0,0.45)$)
      arc[start angle=-90, end angle=75, x radius=0.15, y radius=0.03];
  \end{scope}

  \coordinate (A) at (-0.28,0.45);

  \begin{scope}[rotate around={35:(A)}]
    \draw[->, very thick, red] (A) -- ++(0.17,0);
    \draw[->, very thick, red] (A) -- ++(-0.03,0.13);
    \draw[->, very thick, red] (A) -- ++(0.1,0.14);
  \end{scope}
  \node[text=white] at ($(A) + (-0.12,-0.02)$) {$\{A\}$};

  \coordinate (B) at (0.6,0.75);

  \begin{scope}[rotate around={60:(B)}]
    \draw[->, very thick, red] (B) -- ++(0.2,-0.07);
    \draw[->, very thick, red] (B) -- ++(-0.05,0.18);
    \draw[->, very thick, red] (B) -- ++(-0.12,-0.12);
  \end{scope}
  \node at ($(B) + (0.15,-0.02)$) {$\{B\}$};

  \draw[->, thick, green!70!black] (B) -- ++(-0.2,0.23) 
  	node[left, black] {$v$};

  \draw[->, thick, green!70!black] (B) -- ++(0.09,0.29)
    node[right, xshift=2pt, black] {$\omega_B$};

  \begin{scope}[rotate around={-15:(B)}]
    \draw[->, thick, blue!65!green, transform shape=false] ($(B) + (0.005,0.19)$)
      arc[start angle=-90, end angle=75, x radius=0.11, y radius=0.02];
  \end{scope}

  \begin{scope}[rotate around={-15:(B)},xscale=-1]
    \draw[thick, blue!65!green] ($(B) + (-0.005,0.19)$)
      arc[start angle=-90, end angle=75, x radius=0.11, y radius=0.02];
  \end{scope}

  \draw[black] (O) -- (A)  
  	node[midway, below left, text=black] {$\idx{F}{0}{}{A}$};
  \draw[black] (O) -- (B)
  	node[midway, below right, xshift=1pt, text=black] {$\idx{F}{0}{}{B}$};
  \draw[black] (A) -- (B)
  	node[midway, above, yshift=5pt, text=black] {$\idx{F}{A}{}{B}$};
\end{tikzpicture}

%% file: figures/tikz/earth_geoid_tikz_RAM.tex
\begin{tikzpicture}[scale=4, >=stealth, line cap=round]
  \tikzset{
    com/.style={
      draw,
      circle,
      fill=white,
      minimum size=6pt,
      inner sep=0pt,
      line width=0.6pt,
      path picture={
        \draw[line width=0.6pt]
          (path picture bounding box.south west) --
          (path picture bounding box.north east)
          (path picture bounding box.north west) --
          (path picture bounding box.south east);
      }
    }
  }

  \coordinate (O) at (0,0);
  
  \coordinate (A) at (0.423,0.264);
  
  \node[opacity=0.15] at (O) {\includegraphics[width=4.8cm,height=3cm]%
  	{figures/aussie_globe_nobackground.png}};

  \draw[dashed, thick, black!65] (O) -- ++(0,-0.45);
  \draw[dashed, thick, black!65] (O) -- ++(0, 0.55);

  \draw[->, thick, blue!65!green, transform shape=false] ($(O) + (0,0.45)$) 
    arc[start angle=-90, end angle=75, x radius=0.15, y radius=0.03];

  \begin{scope}[xscale=-1]
    \draw[thick, blue!65!green] ($(O) + (0,0.45)$)
      arc[start angle=-90, end angle=75, x radius=0.15, y radius=0.03];
  \end{scope}
  
  \draw[semithick, black] (O) ellipse [x radius=0.6cm, y radius=0.375cm]
    node[font=\small, left, text=black, align=center, xshift=-2.5cm] {Earth\\ geoid};
  
  \node[com] at (O) {};
  
  \draw[dashed, thick, cyan!80] ($(A) - (0.15,0)$) -- (O);
  \draw[dashed, thick, cyan!80] (O) -- ($(O) + (0.15,0)$);
  
   \draw[->, thick, red] (A) -- ($(A) - (0.15,0)$)
  	node[font=\small, left, text=black, xshift=0.05cm, yshift=0.10cm] 
  	  {$a_A^{\text{c}}$};   
   \draw[->, thick, red] (A) -- ($(O) + (0.15,0)$)
  	node[font=\small, midway, right, text=black, xshift=-0.1cm, yshift=-0.2cm] 
  	  {$g_0 \eb_3$};  
  
  \draw[->, thick, black] (A) -- (O)
  	node[font=\small, left, text=black, xshift=-0.1cm]
  	  {$g_A$};
   \draw[->, thick, black] (A) -- ($(A) + (0.423,0.264) - (0.15,0)$)
  	node[font=\small,  midway, right, text=black, xshift=-0.1cm, yshift=-0.2cm] 
  	  {$a_A^{\text{n}}$};
\end{tikzpicture}

%% file: arxiv_Galilean_20261116.bbl
\begin{thebibliography}{FGvG{\etalchar{+}}25}

\bibitem[Bar24]{barfoot2024state}
Timothy~D Barfoot.
\newblock {\em State estimation for robotics}.
\newblock Cambridge University Press, 2nd edition, 2024.

\bibitem[Bar25]{barfoot2025integral}
Timothy~D Barfoot.
\newblock Integral forms in matrix {Lie} groups.
\newblock {\em arXiv preprint arXiv:2503.02820}, 2025.

\bibitem[BB14]{barrau2014invariant}
Axel Barrau and Silv\`{e}re Bonnabel.
\newblock Invariant particle filtering with application to localization.
\newblock In {\em 53rd IEEE Conference on Decision and Control}, pages 5599--5605. IEEE, 2014.

\bibitem[BB16]{barrau2016invariant}
Axel Barrau and Silvere Bonnabel.
\newblock The invariant extended {Kalman} filter as a stable observer.
\newblock {\em IEEE Transactions on Automatic Control}, 62(4):1797--1812, 2016.

\bibitem[BB20a]{2020_Barrau_Mathematical}
Axel Barrau and Silvere Bonnabel.
\newblock A mathematical framework for {IMU} error propagation with applications to preintegration.
\newblock In {\em 2020 IEEE International Conference on Robotics and Automation (ICRA)}, pages 5732--5738. IEEE, 2020.

\bibitem[BB20b]{barrau2020mathematical}
Axel Barrau and Silvere Bonnabel.
\newblock A mathematical framework for imu error propagation with applications to preintegration.
\newblock In {\em 2020 IEEE International Conference on Robotics and Automation (ICRA)}, pages 5732--5738. IEEE, 2020.

\bibitem[BB23]{Barrau_2023}
Axel Barrau and Silvere Bonnabel.
\newblock The geometry of navigation problems.
\newblock {\em IEEE Transactions on Automatic Control}, 68(2):689–704, February 2023.

\bibitem[BBCB21]{brossard2021associating}
Martin Brossard, Axel Barrau, Paul Chauchat, and Silv{\`e}re Bonnabel.
\newblock Associating uncertainty to extended poses for on {Lie} group {IMU} preintegration with rotating earth.
\newblock {\em IEEE Transactions on Robotics}, 38(2):998--1015, 2021.

\bibitem[BMS09]{2009_bonnable_vaa}
Silvere Bonnable, Philippe Martin, and Erwan Sala{\"u}n.
\newblock Invariant extended kalman filter: theory and application to a velocity-aided attitude estimation problem.
\newblock In {\em Proceedings of the 48h IEEE Conference on Decision and Control (CDC) held jointly with 2009 28th Chinese Control Conference}, pages 1297--1304. IEEE, 2009.

\bibitem[CD18]{2018_Carpenter_navigation}
J~Russell Carpenter and Christopher~N D'Souza.
\newblock Navigation filter best practices.
\newblock Technical Report NASA/TP–2018–219822, NASA, Mail Stop 148, NASA Langley Research Center, Hampton, VA 23681-2199, USA., 2018.
\newblock NASA STI Program.

\bibitem[Chi09]{chirikjian2009stochastic}
Gregory~S Chirikjian.
\newblock {\em Stochastic Models, Information Theory, and Lie Groups, Volume 1: Classical Results and Geometric Methods}.
\newblock Springer Science \& Business Media, 2009.

\bibitem[DFMW24]{Delama_2024}
Giulio Delama, Alessandro Fornasier, Robert Mahony, and Stephan Weiss.
\newblock Equivariant {IMU} preintegration with biases: A {Galilean} group approach.
\newblock {\em IEEE Robotics and Automation Letters}, page 1–8, 2024.

\bibitem[DSG{\etalchar{+}}26]{2026_Delama}
Giulio Delama, Martin Scheiber, Yixiao Ge, Tarek Hamel, Stephan Weiss, and Robert Mahony.
\newblock Galilean state estimation for inertial navigation systems with unknown time delay.
\newblock In {\em Proceedings of Robotics Science and Systems}, 2026.

\bibitem[FAM{\etalchar{+}}19]{2019_Fourmy_Absolute}
Mederic Fourmy, Dinesh Atchuthan, Nicolas Mansard, Joan Sol\`{a}, and Thomas Flayols.
\newblock Absolute humanoid localization and mapping based on {IMU} {Lie} group and fiducial markers.
\newblock In {\em Proceedings of the {IEEE} International Conference on Humanoid Robots}, pages 237--243, Toronto, Ontario, Canada, October 2019.

\bibitem[FCDS16]{forster2016manifold}
Christian Forster, Luca Carlone, Frank Dellaert, and Davide Scaramuzza.
\newblock On-manifold preintegration for real-time visual--inertial odometry.
\newblock {\em IEEE transactions on robotics}, 33(1):1--21, 2016.

\bibitem[FGvG{\etalchar{+}}25]{FornasierEquivariantSymmetries2025}
Alessandro Fornasier, Yixiao Ge, Pieter van Goor, Robert Mahony, and Stephan Weiss.
\newblock Equivariant symmetries for inertial navigation systems.
\newblock {\em Automatica}, 181:112495, 2025.

\bibitem[Gie21]{2021_Giefer_Uncertainties}
Lino~Antoni Giefer.
\newblock Uncertainties in {Galilean} spacetime.
\newblock In {\em Proceedings of the {IEEE} International Conference on Information Fusion {(FUSION)}}, Sun City, South Africa, November 2021.

\bibitem[GMD20]{gill2020full}
Rajan Gill, Mark~W Mueller, and Raffaello D’Andrea.
\newblock Full-order solution to the attitude reset problem for {Kalman} filtering of attitudes.
\newblock {\em Journal of Guidance, Control, and Dynamics}, 43(7):1232--1246, 2020.

\bibitem[GPSA02]{2002_Goldstein_mechanics}
Herbert Goldstein, Charles Poole, John Safko, and Stephen~R Addison.
\newblock {\em Classical mechanics}.
\newblock American Association of Physics Teachers, 2002.

\bibitem[GvGM25]{ge2025geometryextendedkalmanfilters}
Yixiao Ge, Pieter van Goor, and Robert Mahony.
\newblock The geometry of extended {Kalman} filters on manifolds with affine connection, 2025.

\bibitem[GvM24]{2024_Ge_LCSS}
Yixiao Ge, Pieter {van Goor}, and Robert Mahony.
\newblock A geometric perspective on fusing gaussian distributions on {Lie} groups.
\newblock {\em IEEE Control Systems Letters}, 8:844--849, May 2024.
\newblock Presented at the 63rd Conference on Decision and Control (CDC2024).

\bibitem[Hol11]{2011_Holm_Geometric_Part_II}
Darryl~D Holm.
\newblock {\em Geometric mechanics-part II: Rotating, translating and rolling}.
\newblock World Scientific, 2011.

\bibitem[Kel23]{2024_Kelly_arxiv}
Jonathan Kelly.
\newblock All about the galilean group {SGal(3)}.
\newblock Technical Report STARS-2023-001, University of Toronto Institute for Aerospace Studies, Toronto, Ontario, Canada, 2023.

\bibitem[Ken82]{kent1982fisher}
John~T Kent.
\newblock The fisher-bingham distribution on the sphere.
\newblock {\em Journal of the Royal Statistical Society: Series B (Methodological)}, 44(1):71--80, 1982.

\bibitem[KGG21]{kelly2021question}
Jonathan Kelly, Christopher Grebe, and Matthew Giamou.
\newblock A question of time: Revisiting the use of recursive filtering for temporal calibration of multisensor systems.
\newblock In {\em 2021 IEEE International Conference on Multisensor Fusion and Integration for Intelligent Systems (MFI)}, pages 1--8. IEEE, 2021.

\bibitem[KS14]{kelly2014general}
Jonathan Kelly and Gaurav~S Sukhatme.
\newblock A general framework for temporal calibration of multiple proprioceptive and exteroceptive sensors.
\newblock In {\em Experimental Robotics: The 12th International Symposium on Experimental Robotics}, pages 195--209. Springer, 2014.

\bibitem[LL71]{levy1971galilei}
Jean-Marc L{\'e}vy-Leblond.
\newblock {Galilei} group and {Galilean} invariance.
\newblock In {\em Group theory and its applications}, pages 221--299. Elsevier, 1971.

\bibitem[LS11]{lupton2011visual}
Todd Lupton and Salah Sukkarieh.
\newblock Visual-inertial-aided navigation for high-dynamic motion in built environments without initial conditions.
\newblock {\em IEEE Transactions on Robotics}, 28(1):61--76, 2011.

\bibitem[Mah24]{2024_Mahony_talk}
Robert Mahony.
\newblock {Galilean} space-time in robotics.
\newblock In {\em Equivariant Robotics: The Role of Symmetry Across Perception, Estimation, and Control; at IEEE/RSJ International Conference on Intelligent Robots and Systems}, 2024.
\newblock Keynote.

\bibitem[Mak67]{maksimov1967necessary}
Vyacheslav~Mikhailovich Maksimov.
\newblock Necessary and sufficient statistics for the family of shifts of probability distributions on continuous bicompact groups.
\newblock {\em Theory of Probability \& Its Applications}, 12(2):267--280, 1967.

\bibitem[Mar03]{markley2003attitude}
F~Landis Markley.
\newblock Attitude error representations for {Kalman} filtering.
\newblock {\em Journal of guidance, control, and dynamics}, 26(2):311--317, 2003.

\bibitem[May82]{1982_Maybeck_stochastic_book}
Peter~S Maybeck.
\newblock {\em Stochastic models, estimation, and control}, volume~3.
\newblock Academic press, 1982.

\bibitem[MHD17]{mueller2017covariance}
Mark~W Mueller, Markus Hehn, and Raffaello D’Andrea.
\newblock Covariance correction step for {Kalman} filtering with an attitude.
\newblock {\em Journal of Guidance, Control, and Dynamics}, 40(9):2301--2306, 2017.

\bibitem[MHT21]{2021_mahony_HomogeneousSpaceGeometry}
Robert Mahony, Tarek Hamel, and Jochen Trumpf.
\newblock An homogeneous space geometry for simultaneous localisation and mapping.
\newblock {\em Annual Reviews in Control}, June 2021.

\bibitem[MR99]{1999_Marsden_Introduction}
Jerrold~E. Marsden and Tudor~S. Ratiu.
\newblock {\em Introduction to Mechanics and Symmetry: A Basic Exposition of Classical Mechanical Systems}, volume~17 of {\em Texts in Applied Mathematics}.
\newblock Springer, New York, New York, USA, 2 edition, 1999.

\bibitem[PHKF12]{pavlis2012development}
Nikolaos~K Pavlis, Simon~A Holmes, Steve~C Kenyon, and John~K Factor.
\newblock The development and evaluation of the earth gravitational model 2008 (egm2008).
\newblock {\em Journal of geophysical research: solid earth}, 117(B4), 2012.

\bibitem[RSB99]{1999_Roumeliotis_errorstate}
Stergios~I Roumeliotis, Gaurav~S Sukhatme, and George~A Bekey.
\newblock Circumventing dynamic modeling: Evaluation of the error-state kalman filter applied to mobile robot localization.
\newblock In {\em Proceedings 1999 IEEE International Conference on Robotics and Automation (Cat. No. 99CH36288C)}, volume~2, pages 1656--1663. IEEE, 1999.

\bibitem[SCNF24]{2024_Shalaby_Multi-robot}
Mohammed~Ayman Shalaby, Charles~Champagne Cossette, Jerome~Le Ny, and James~Richard Forbes.
\newblock Multi-robot relative pose estimation and {IMU} preintegration using passive {UWB} transceivers.
\newblock {\em {IEEE} Transactions on Robotics}, 40:2410--2429, 2024.

\bibitem[SDA21]{2021_Sola_Micro}
Joan Sol\`{a}, Jeremie Deray, and Dinesh Atchuthan.
\newblock A micro {Lie} theory for state estimation in robotics.
\newblock {\em arXiv e-prints}, 2021.

\bibitem[Sel05]{2005_Selig_Geometric}
Jon~M. Selig.
\newblock {\em Geometric Fundamentals of Robotics}.
\newblock Monographs in Computer Science. Springer-Verlag, New York, 2 edition, 2005.

\bibitem[vGM21]{van2021autonomous}
Pieter van Goor and Robert Mahony.
\newblock Autonomous error and constructive observer design for group affine systems.
\newblock In {\em 2021 60th IEEE Conference on Decision and Control (CDC)}, pages 4730--4737. IEEE, 2021.

\bibitem[vHM23]{2023_vanGoor_EqF}
Pieter {van Goor}, Tarek Hamel, and Robert Mahony.
\newblock Equivariant filter ({EqF}).
\newblock {\em IEEE Transactions on Automatic Control}, 68(6):3501--3512, 2023.

\bibitem[Wat84]{watson1984theory}
Geoffrey~S Watson.
\newblock The theory of concentrated langevin distributions.
\newblock {\em Journal of multivariate analysis}, 14(1):74--82, 1984.

\bibitem[WC06]{Wang_2006_Error}
Yunfeng Wang and G.S. Chirikjian.
\newblock Error propagation on the {{Euclidean}} group with applications to manipulator kinematics.
\newblock {\em IEEE Transactions on Robotics}, 22(4):591--602, August 2006.

\end{thebibliography}
